\documentclass{article} 
\usepackage[preprint]{nips_2018}
\usepackage[utf8]{inputenc} 
\usepackage[T1]{fontenc}    
\usepackage{xcite}
\usepackage{xr}
\usepackage{xr-hyper} 
\usepackage{hyperref}       
\usepackage{url}            
\usepackage{booktabs}       
\usepackage{amsfonts}       
\usepackage{nicefrac}       
\usepackage{microtype}      
\usepackage{graphicx}
\usepackage{subcaption}

\usepackage{amsmath}
\usepackage{bm}
\usepackage{amsthm}
\usepackage{multirow}
\usepackage{algorithm}
\usepackage[noend]{algpseudocode}
\usepackage{caption}
\usepackage[subtle]{savetrees}

\title{Adversarially Learned Mixture Model}

\author{
  Andrew Jesson \\
  Imagia Cybernetics Inc.\\
  Montreal, QC, Canada \\
  \texttt{andrew.jesson@imagia.com} \\
  \And
  C\'ecile Low-Kam \\
  Imagia Cybernetics Inc. \\
  Montreal, QC, Canada \\
  \texttt{cecile.low-kam@imagia.com} \\
  \AND
  Florian Soudan \\
  Imagia Cybernetics Inc. \\
  Montreal, QC, Canada \\
  \texttt{florian@imagia.com} \\
  \And
  Nicolas Chapados \\
  Imagia Cybernetics Inc. \\
  Montreal, QC, Canada \\
  \texttt{nic@imagia.com} \\
}

\begin{document}
\maketitle

\begin{abstract}
The Adversarially Learned Mixture Model (AMM) is a generative model for unsupervised or semi-supervised data clustering. The AMM is the first adversarially optimized method to model the conditional dependence between inferred continuous and categorical latent variables. Experiments on the MNIST and SVHN datasets show that the AMM allows for semantic separation of complex data when little or no labeled data is available. The AMM achieves a state-of-the-art unsupervised clustering error rate of 2.86\% on the MNIST dataset. A semi-supervised extension of the AMM yields competitive results on the SVHN dataset. 
\end{abstract}

\section{Introduction}

Semi-supervised or unsupervised representation learning enables the utilization of all available data when tackling problems where there are little or no labeled examples. This is a common scenario in many applications of machine learning, such as medical image analysis, where it is reinforced by the expense of obtaining expert labeled examples. Moreover, machine-learned representations are more likely to be used for subsequent tasks if they are interpretable and meaningful. Deep generative modelling is a suitable approach to this problem, as derived models have been shown capable of learning from both labeled and unlabeled examples, embedding data according to desired latent variable distributions, and producing realistic data examples generated from samples of those latent variables.

The Generative Adversarial Network (GAN) has recently emerged as a powerful framework for modeling complex data distributions without having to approximate intractable likelihoods. In the formulation by \citet{goodfellow2014generative}, a GAN consists of two networks: a \emph{generator} $G$ that is trained to yield unique samples from the data distribution,
and a \emph{discriminator} $D$ that is trained to distinguish between generated and true data samples.

\citet{dumoulin2016adversarially} and \citet{donahue2016adversarial} have proposed the ALI and BiGAN models that add an inference process, i.e., the ability to map data samples to points in the latent space, to the GAN framework. A second generator for inference, or \emph{encoder}, is added to the original GAN generator and the discriminator is adapted for the two-dimensional space of data inputs and latent representations. A variant of the resulting model is also introduced by \citet{dumoulin2016adversarially} for conditional data generation, but still assumes that the class of the data is always observed, as inference of categorical variables is not included.

Adversarial approaches for the inference of both continuous and categorical variables are actively researched. \citet{chen2016infogan} introduce a hybrid adversarial method that is capable of modelling both continuous and categorical latent variables for unsupervised clustering and feature disentanglement. Another hybrid adversarial method is introduced by \citet{makhzani2015adversarial} where adversarial objectives on continuous and categorical latent variables are optimized for unlabeled examples and categorical cross entropy on categorical variables is optimized for labeled examples. \citet{li2017triplegan} and \citet{deng2017structured} point toward fully adversarial semi-supervised classification using inferred categorical variables by introducing a ``three player'' adversarial game, but stop short by adding auxiliary ``collaborative'' objectives. In each of these methods, it is assumed that categorical and continuous latent variables are independently distributed. This independence assumption results in discontinuities in the latent space between categories, which removes the notion of inter-categorical proximity.

Another notable family of generative models, Variational Autoencoders (VAEs), maximize the posterior distribution of latent representations given the data instead of using an adversarial approach. As VAEs integrate inference, semi-supervised classification can be performed by conditioning the continuous latent variable of the VAE on the class label \citep{kingma2014semi, dilokthanakul2016deep,maaloe2017semi}. However, the quality of VAE results depend on the expressiveness of the inference distribution and every time the assumptions about the inference or data distributions are changed a new objective function needs to be derived. In this way, variational optimization is not as versatile as adversarial training.

We present the Adversarially Learned Mixture Model (AMM). The AMM is, to our knowledge, the first generative model inferring both continuous and categorical latent variables to perform either unsupervised or semi-supervised clustering of data using a single adversarial objective. This is enabled, in part, by explicitly modelling the dependence between continuous and categorical latent variables, which eliminates discontinuities between categories in the latent space. Semi-supervised clustering and classification is enabled by a simplified formulation of the ``three player game'', presented by \citet{li2017triplegan}. In this paper we show that the AMM achieves state of the art unsupervised clustering error rate on the MNIST dataset \citep{lecun-mnisthandwrittendigit-2010}, and that it achieves competitive results for semi-supervised classification on the SVHN dataset \citep{netzer2011svhn}.

\section{Method}
\label{sec:method}
\subsection{Preliminaries}
The ALI and BiGAN models are trained by matching two joint distributions of images \(\bm{x} \in \mathbb{R}^D\) and their latent code \(\bm{z} \in \mathbb{R}^L\). The two distributions to be matched are the inference distribution \(q(\bm{x}, \bm{z})\) and the synthesis distribution \(p(\bm{x}, \bm{z})\), where,
\begin{eqnarray}
    q(\bm{x},\bm{z}) & = & q(\bm{x})q(\bm{z} \mid \bm{x}),
\\
    p(\bm{x},\bm{z}) & = & p(\bm{z})p(\bm{x} \mid \bm{z}).
\end{eqnarray}
Samples of \(q(\bm{x})\) are drawn from the training data and samples of \(p(\bm{z})\) are drawn from a prior distribution, usually \(\mathcal{N}(0, 1)\). Samples from \(q(\bm{z} \mid \bm{x})\) and \(p(\bm{x} \mid \bm{z})\) are drawn from neural networks that are optimized during training. \citet{dumoulin2016adversarially} show that sampling from \(q(\bm{z} \mid \bm{x}) = \mathcal{N}(\mu(\bm{x}), \sigma^2(\bm{x})I)\) is possible by employing the reparametrization trick \citep{kingma13autoencodingVB}, i.e.\ computing
\begin{equation}
    \bm{z} = \mu(\bm{x}) + \sigma(\bm{x}) \odot \epsilon, \quad \epsilon \sim \mathcal{N}(0, I),
\end{equation}

where $\odot$ is element wise vector multiplication.

A conditional variant of ALI has also been explored by \citet{dumoulin2016adversarially} where an observed class-conditional categorical variable \(\bm{y}\) has been introduced. The joint factorization of each distribution to be matched are:
\begin{eqnarray}
    q(\bm{x},\bm{y},\bm{z}) & = &  q(\bm{x},\bm{y})q(\bm{z} \mid \bm{y},\bm{x}),
    \label{eq:supervised_inference}
\\
    p(\bm{x},\bm{y},\bm{z}) & = & p(\bm{y})p(\bm{z})p(\bm{x} \mid \bm{y},\bm{z}).
\end{eqnarray}
Samples of \(q(\bm{x},\bm{y})\) are drawn from the data. Samples of \(p(\bm{z})\) are drawn from a continuous prior on \(\bm{z}\), and samples of \(p(\bm{y})\) are drawn from a categorical prior on \(\bm{y}\), both of which are marginally independent. Samples from \(q(\bm{z} \mid \bm{y},\bm{x})\) and \(p(\bm{x} \mid \bm{y},\bm{z})\) are drawn from neural networks that are optimized during training. 

In the following sections we present graphical models for \(q(\bm{x}, \bm{y}, \bm{z})\) and \(p(\bm{x}, \bm{y}, \bm{z})\) that build off of conditional ALI. Where conditional ALI requires the full observation of categorical variables, the models we present will account for both unobserved and partially observed categorical variables. We finally show how they can be optimized using a single adversarial objective.

\subsection{Adversarially Learned Mixture Model}

\begin{figure}
\centering 
\includegraphics[width=0.5\textwidth]{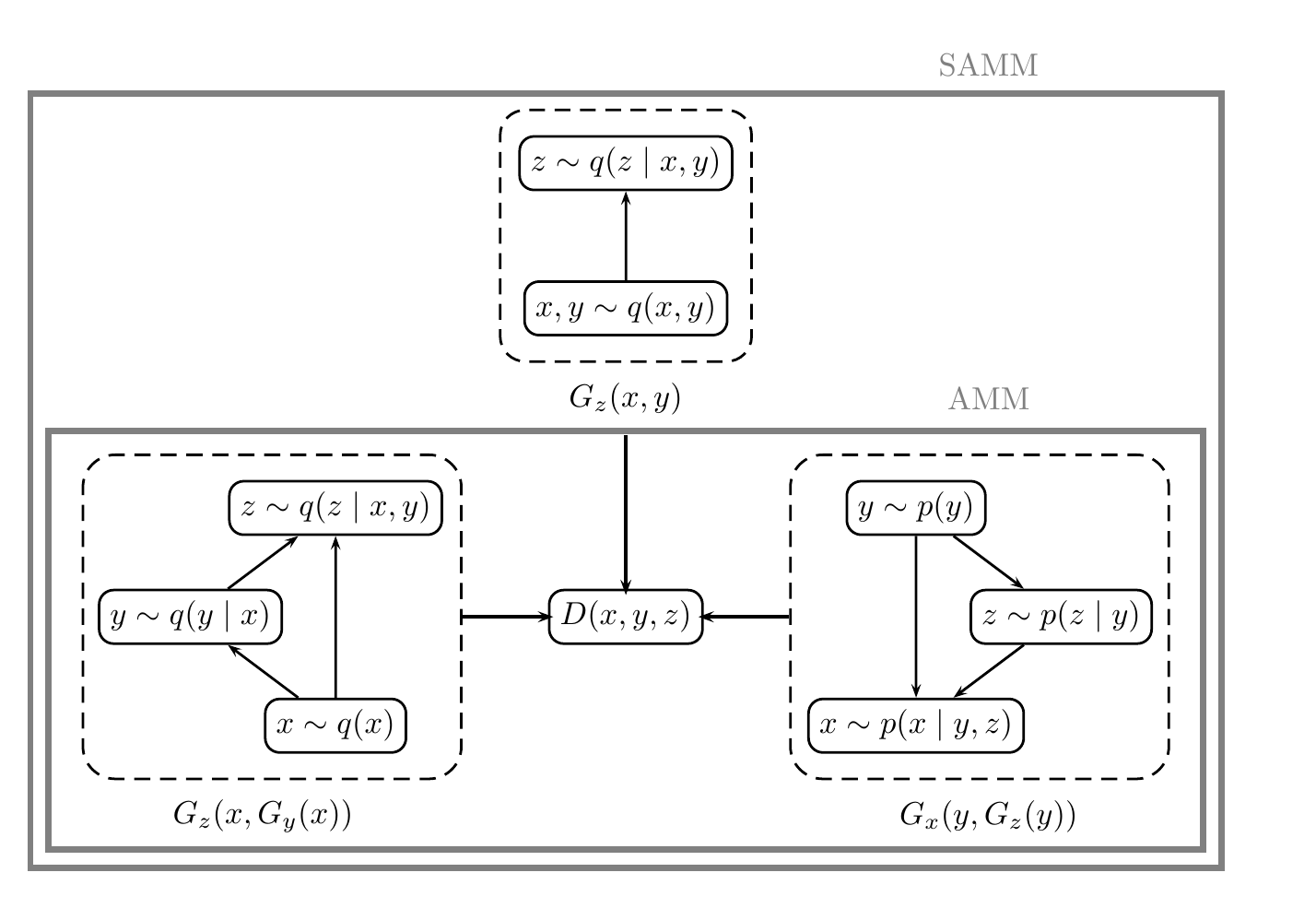}
\caption{Overview of the unsupervised (AMM) and semi-supervised (SAMM) model with the first option (Equation \eqref{eq:unsupervised_inference_1}) for the inference distribution. AMM consists of two generators, encoder $G_{\bm{z}}(\bm{x},G_{\bm{y}}(\bm{x}))$ and decoder $G_{\bm{x}}(\bm{y},\bm{z})$, and a discriminator $D(\bm{x},\bm{y},\bm{z})$. SAMM includes an additional generator for labeled data, $G_{z}(\bm{x},\bm{y})$.}
\label{fig:model_overview}
\end{figure}

The AMM is an adversarial generative model for deep unsupervised clustering of data.  Figure \ref{fig:model_overview} presents an overview of the model.

Like conditional ALI, a categorical variable is introduced to model the labels. However, the unsupervised setting now requires a different factorization of the inference distribution in order to enable inference of the categorical variable $\bm{y}$, namely:
\begin{equation}
    q_1(\bm{x},\bm{y},\bm{z}) = q(\bm{x})q(\bm{y} \mid \bm{x})q(\bm{z} \mid \bm{x},\bm{y}),
    \label{eq:unsupervised_inference_1}
\end{equation}
or
\begin{equation}
    q_2(\bm{x},\bm{y},\bm{z}) = q(\bm{x})q(\bm{z} \mid \bm{x})q(\bm{y} \mid \bm{x},\bm{z}).
    \label{eq:unsupervised_inference_2}
\end{equation}
Samples of \(q(\bm{x})\) are drawn from the training data, and samples from \(q(\bm{y} \mid \bm{x})\), \(q(\bm{z} \mid \bm{x},\bm{y})\) or \(q(\bm{z} \mid \bm{x})\), \(q(\bm{y} \mid \bm{x},\bm{z})\) are generated by neural networks. 
We follow
\citet{kendall2017uncertainties} and sample from \(q(\bm{y} \mid \bm{x})\) by computing
\begin{eqnarray}
    h_{\bm{y}}(\bm{x}) & = & \mu_{\bm{y}}(\bm{x}) + \sigma_{\bm{y}}(\bm{x}) \odot \epsilon, \quad \epsilon \sim \mathcal{N}(0, I),
\\
    y(\bm{x}) & = & \textrm{softmax}\left(h_{\bm{y}}(\bm{x})\right).
\label{eq:y}
\end{eqnarray}
Then, we can sample from \(q(\bm{z} \mid \bm{x},\bm{y})\) by computing
\begin{equation}
\begin{split}
    z(\bm{x}, h_{\bm{y}}(\bm{x})) = \mu_{\bm{z}}(\bm{x}, h_{\bm{y}}(\bm{x})) + \sigma_{\bm{z}}(\bm{x}, h_{\bm{y}}(\bm{x})) \odot \epsilon, \quad \epsilon \sim \mathcal{N}(0, I).
    \end{split}
\label{eq:z}
\end{equation}
A similar sampling strategy can be used to sample from \(q(\bm{y} \mid \bm{x},\bm{z})\) in \eqref{eq:unsupervised_inference_2}.

The factorization of the synthesis distribution \(p(\bm{x}, \bm{y}, \bm{z})\) also differs from conditional ALI:
\begin{equation}
    p(\bm{x},\bm{y},\bm{z}) = p(\bm{y})p(\bm{z} \mid \bm{y})p(\bm{x} \mid \bm{y},\bm{z}).
    \label{eq:unsupervised_synthesis}
\end{equation}
The product \(p(\bm{y})p(\bm{z} \mid \bm{y})\) can be conveniently given by a mixture model. Samples from \(p(\bm{y})\) are drawn from a multinomial prior, and samples from \(p(\bm{z} \mid \bm{y})\) are drawn from a continuous prior, for example, \(\mathcal{N}(\mu_{\bm{y}=k},1)\). Samples from \(p(\bm{z} \mid \bm{y})\) can alternatively be generated by a neural network by again employing the reparameterization trick. Namely,
\begin{equation}
\begin{split}
    z(\bm{y}) = \mu(\bm{y}) + \sigma(\bm{y}) \odot \epsilon, \quad \epsilon \sim \mathcal{N}(0, I).
    \end{split}
\label{eq:z_learned}
\end{equation}
This approach effectively learns the parameters of \(\mathcal{N}(\mu_{\bm{y}=k},\sigma_{\bm{y}=k})\).

\subsubsection{Adversarial Value Function}
We follow \citet{dumoulin2016adversarially} and define the value function that describes the unsupervised game between the discriminator $D$ and the generator $G$ as:
\begin{equation}
\label{eq:unsupervised_value_function}
\begin{split}
    \min_G \max_D V(D, G) &= \mathbb{E}_{q(\bm{x})}[\log(D(\bm{x}, G_{\bm{y}}(\bm{x}), G_{\bm{z}}(\bm{x}, G_{\bm{y}}(\bm{x}))))] \\
	                      &\quad + \mathbb{E}_{p(\bm{y}, \bm{z})}[\log(1 - D(G_{\bm{x}}(\bm{y}, G_{\bm{z}}(\bm{y})), \bm{y}, G_{\bm{z}}(\bm{y})))] \\
                          &= \iiint q(\bm{x}) q(\bm{y} \mid \bm{x}) q(\bm{z} \mid \bm{x}, \bm{y}) \log(D(\bm{x}, \bm{y}, \bm{z})) \,d\bm{x} \,d\bm{y} \,d\bm{z} \\
                          &\quad + \iiint p(\bm{y}) p(\bm{z} \mid \bm{y}) p(\bm{x} \mid \bm{y,} \bm{z}) \log(1 - D(\bm{x}, \bm{y}, \bm{z})) \,d\bm{x} \,d\bm{y} \,d\bm{z}.
\end{split}
\end{equation}

There are four generators in total: two for the encoder $G_{\bm{y}}(\bm{x})$ and $G_{\bm{z}}(\bm{x},G_{\bm{y}}(\bm{x}))$, which map the data samples to the latent space; and two for the decoder $G_{\bm{z}}(\bm{y})$ and $G_{\bm{x}}(\bm{y},G_{\bm{z}}(\bm{y}))$, which map samples from the prior to the input space. $G_{\bm{z}}(\bm{y})$ can either be a learned function, or be specified by a known prior. See Algorithm \ref{alg:amm} for a detailed description of the optimization procedure.

\begin{algorithm*}[!h]
    \begin{algorithmic}
        \State $\theta_{G_{\bm{y}}(\bm{x})}, \theta_{G_{\bm{z}}(\bm{x}, G_{\bm{y}}(\bm{x}))}, \theta_{G_{\bm{z}}(\bm{y})}, \theta_{G_{\bm{x}}(\bm{y}, G_{\bm{z}}(\bm{y}))}, \theta_{D}$ \Comment{Initialize AMM parameters}
        \While{not done}
            \State $\bm{x}^{(1)}, \ldots, \bm{x}^{(M)} \sim q(\bm{x})$ \Comment{Sample from data and priors}
            \State $\bm{y}^{(1)}, \ldots, \bm{y}^{(M)} \sim p(\bm{y})$
            \State $\bm{z}^{(j)} \sim p(\bm{z} \mid \bm{y} = \bm{y}^{(j)}), \quad j = 1, \ldots, M$
            
            \State $\tilde{\bm{x}}^{(j)} \sim p(\bm{x} \mid \bm{y} = \bm{y}^{(j)}, \bm{z} = \bm{z}^{(j)}), \quad j = 1, \ldots, M$ \Comment{Sample from conditionals}
            \State $\tilde{\bm{y}}^{(i)} \sim q(\bm{y} \mid \bm{x} = \bm{x}^{(i)}), \quad i = 1, \ldots, M$
            \State $\tilde{\bm{z}}^{(i)} \sim q(\bm{z} \mid \bm{x} = \bm{x}^{(i)}, \bm{y} = \tilde{\bm{y}}^{(i)}), \quad i = 1, \ldots, M$
            
            \State $\rho_{q}^{(i)} \gets D(\bm{x}^{(i)}, \tilde{\bm{y}}^{(i)}, \tilde{\bm{z}}^{(i)}), \quad i = 1, \ldots, M$  \Comment{Compute discriminator predictions}
            \State $\rho_{p}^{(j)} \gets D(\tilde{\bm{x}}^{(j)}, \bm{y}^{(j)}, \bm{z}^{(j)}), \quad j = 1, \ldots, M$
            
            \State $\mathcal{L}_D \gets -\frac{1}{M} \sum_{i=1}^M \log(\rho_q^{(i)}) -\frac{1}{M} \sum_{j=1}^M\ log(1 - \rho_p^{(j)})$ \Comment{Compute discriminator losses}
            \State $\mathcal{L}_{G_{\bm{x}}(\bm{y}, G_{\bm{z}}(\bm{y}))} = \mathcal{L}_{G_{\bm{z}}(\bm{y})} \gets -\frac{1}{M} \sum_{i=1}^M \log(\rho_p^{(i)})$ \Comment{Compute x generator losses}
            \State $\mathcal{L}_{G_{\bm{y}}(\bm{x})} = \mathcal{L}_{G_{\bm{z}}(\bm{x}, G_{\bm{y}}(\bm{x}))} \gets -\frac{1}{M} \sum_{i=1}^M \log(1-\rho_q^{(i)})$ \Comment{Compute y and z generator loss}
            
            \State $\theta_{D} \gets \theta_D - \nabla_{\theta_D} \mathcal{L}_D$ \Comment{Update discriminator parameters}
            \State $\theta_{G_{\bm{x}}(\bm{y}, G_{\bm{z}}(\bm{y}))} \gets \theta_{G_{\bm{x}}(\bm{y}, G_{\bm{z}}(\bm{y}))} - \nabla_{\theta_{G_{\bm{x}}(\bm{y}, G_{\bm{z}}(\bm{y}))}} \mathcal{L}_{G_{\bm{x}}(\bm{y}, G_{\bm{z}}(\bm{y}))}$ \Comment{Update generator parameters}
            \State $\theta_{G_{\bm{z}}(\bm{y})} \gets \theta_{G_{\bm{z}}(\bm{y})} - \nabla_{\theta_{G_{\bm{z}}(\bm{y})}} \mathcal{L}_{G_{\bm{z}}(\bm{y})}$
            \State $\theta_{G_{\bm{y}}(\bm{x})} \gets \theta_{G_{\bm{y}}(\bm{x})} - \nabla_{\theta_{G_{\bm{y}}(\bm{x})}} \mathcal{L}_{G_{\bm{y}}(\bm{x})}$
            \State $\theta_{G_{\bm{z}}(\bm{x}, G_{\bm{y}}(\bm{x}))} \gets \theta_{G_{\bm{z}}(\bm{x}, G_{\bm{y}}(\bm{x}))} - \nabla_{\theta_{G_{\bm{z}}(\bm{x}, G_{\bm{y}}(\bm{x}))}} \mathcal{L}_{G_{\bm{z}}(\bm{x}, G_{\bm{y}}(\bm{x}))}$
        \EndWhile
    \end{algorithmic}
\caption{\label{alg:amm} AMM training procedure using distributions \eqref{eq:unsupervised_inference_1} and \eqref{eq:unsupervised_synthesis}.}
\end{algorithm*}

\subsection{Semi-Supervised Adversarially Learned Mixture Model}

The Semi-Supervised Adversarially Learned Mixture Model (SAMM) is an adversarial generative model for supervised or semi-supervised clustering and classification of data. The objective for training SAMM involves two adversarial games to match pairs of joint distributions. The supervised game matches inference distribution \eqref{eq:supervised_inference} to synthesis distribution \eqref{eq:unsupervised_synthesis} and is described by the following value function:
\begin{equation}
\label{eq:supervised_value_function}
\begin{split}
    \min_G \max_D V(D, G) &= \mathbb{E}_{q(\bm{x}, \bm{y})}[\log(D(\bm{x}, \bm{y}, G_{\bm{z}}(\bm{x}, \bm{y})))] + \mathbb{E}_{p(\bm{y}, \bm{z})}[\log(1 - D(G_{\bm{x}}(\bm{y}, G_{\bm{z}}(\bm{y})), \bm{y}, G_{\bm{z}}(\bm{y})))] \\
                          &= \iiint q(\bm{x}, \bm{y}) q(\bm{z} \mid \bm{x}, \bm{y}) \log(D(\bm{x}, \bm{y}, \bm{z})) \,d\bm{x} \,d\bm{y} \,d\bm{z} \\
                          &\quad + \iiint p(\bm{y}) p(\bm{z} \mid \bm{y}) p(\bm{x} \mid \bm{y,} \bm{z}) \log(1 - D(\bm{x}, \bm{y}, \bm{z})) \,d\bm{x} \,d\bm{y} \,d\bm{z}.
\end{split}
\end{equation}

The unsupervised game matches either of the inference distributions, \eqref{eq:unsupervised_inference_1} or \eqref{eq:unsupervised_inference_2} to the synthesis distribution \eqref{eq:unsupervised_synthesis}. In the case using distribution \eqref{eq:unsupervised_inference_1}, the unsupervised game is described by \eqref{eq:unsupervised_value_function}. The generator for semi-supervised learning has three components: encoders $G_{\bm{z}}(\bm{x},G_{\bm{y}}(\bm{x}))$ and $G_{\bm{z}}(\bm{x},\bm{y})$  map the labeled and unlabeled data samples, respectively, to the latent space, and a decoder $G_{\bm{x}}(\bm{y},G_{\bm{z}}(\bm{y}))$ maps samples of $\bm{y}$ and $\bm{z}$ to the input space, where $G_{\bm{z}}(\bm{z})$ can either be a learned function or be specified by a prior. The encoder for labeled data again consists of two generators (Figure \ref{fig:model_overview}). A detailed description of the training algorithm is given in algorithm \ref{alg:samm} of the appendix. In practice, optimization of each of the generators and the discriminator can be done simultaneously for both the unsupervised and semi-supervised updates.

\section{Related Works}
\label{sec:related}

Unsupervised clustering using hybrid adversarial approaches are proposed by both \citet{makhzani2015adversarial} (AAE) and \citet{chen2016infogan} (InfoGAN). For AAE, the synthesis generator is optimized by minimizing the per-example L2 loss between between input data $\{\bm{x}_i\}$ and their reconstructions $\{\dot{\bm{x}}_i = G_{\bm{x}_i}(G_{\bm{y}}(\bm{x}_i), G_{\bm{z}}(\bm{x}_i))$\}, while the inference generator is optimized using both the L2 objective and an adversarial objective. For InfoGAN, the inference generator is optimized by maximizing the per-example Mutual Information (MI) between samples of categorical latent variables $\{\bm{y}_i \sim p(\bm{y})\}$ and continuous latent variables $\{\bm{z}_i \sim p(\bm{z})\}$ and their ``reconstructions'' $\{\{\dot{\bm{y}}_i, \dot{\bm{z}}_i\}  = G_{\bm{y}, \bm{z}}(G_{\bm{x}}(\bm{y}_i, \bm{z}_i)$\}, while the synthesis generator is optimized using both the MI objective and an adversarial objective. 

On the other end of the generative spectrum, \citet{dilokthanakul2016deep} and \citet{jiang2016variational} offer non-adversarial, VAE-based approaches for unsupervised clustering. Like in the AMM, the combination of priors for the latent variables $\bm{y}$ and $\bm{z}$ is modeled as a Gaussian mixture model, where $\bm{y}$ corresponds to the mixture components.

Multiple adversarial methodologies have been proposed for supervised or semi-supervised learning \citep{springenberg2015unsupervised, salimans2016improved, miyato2017virtual}, but they suffer from the same limitation as the original GAN: they do not provide inference. \citet{gan2017trianglegan}, \citet{li2017triplegan} and \citet{deng2017structured} introduce a third player to the adversarial game. Although this extra player allows to infer categorical variables, these approaches are not fully adversarial as auxiliary ``collaborative'' terms are added to the objective function. Moreover, categorical and continuous latent variables are modeled independently. 

The adversarial and hybrid-adversarial approaches thus far discussed all model $\bm{y}$ and $\bm{z}$ as being conditionally independent from each other. This may be an ideal prior structure for inference, for example, in learning disentangled representations of $\bm{x}$ sampled from a limited domain \citep{chen2016infogan}. However, the independence assumption cannot account for the notion of proximity between categories because $\bm{z}$ is identically distributed for each category in $\bm{y}$. Therefore, the distance between categories is equal and indeterminate. AMM and SAMM are presented as adversarial approaches to model conditional dependencies between $\bm{y}$ and $\bm{z}$, but they do not preclude the independence assumption. The proposed methods can model $\bm{y}$ and $\bm{z}$ as conditionally independent with inference distribution
\begin{equation}
    q(\bm{x},\bm{y},\bm{z}) = q(\bm{x})q(\bm{y} \mid \bm{x})q(\bm{z} \mid \bm{x}),
    \label{eq:independent_unsupervised_inference}
\end{equation}
and synthesis distribution
\begin{equation}
    p(\bm{x},\bm{y},\bm{z}) = p(\bm{y})p(\bm{z})p(\bm{x} \mid \bm{y},\bm{z});
    \label{eq:independent_unsupervised_synthesis}
\end{equation}

however, analysis of this graphical model is left for future work.

\section{Evaluation}
\label{sec:evaluation}

AMM and SAMM are evaluated using two image datasets: MNIST \citep{lecun-mnisthandwrittendigit-2010} and SVHN \citep{netzer2011svhn}. The provided training and testing splits are used for MNIST experiments with 5000 randomly selected examples left out of the training set for validation. The same training, testing, and validation splits as \citet{dumoulin2016adversarially} are used for SVHN. Preprocessing is limited to scaling image intensities on the range $[0, 1]$. Detailed architectures for each experiment are shown in figure \ref{fig:amm_arch} of the appendix. We optimize all networks using Adam \citep{kingma2014adam} with $\alpha = 0.0002$ and $\beta_1 = 0.5$. All kernel weights are initialized using a Gaussian distribution with standard deviation 0.02, all biases are initialized to 0.0.

\subsection{Gradient Penalty}
The gradient penalty introduced by \citet{gulrajani2017improved} is added to the discriminator loss to help stabilize training of AMM and SAMM models. This penalty keeps the gradients of the discriminator with respect to the inputs \(\bm{x}\), \(\bm{y}\), and \(\bm{z}\) on the same order of magnitude. The penalty applied to the discriminator loss is 
\begin{equation}
    \mathcal{L}_{\nabla_{\hat{\bm{x}}, \hat{\bm{y}}, \hat{\bm{z}}}} = \lambda \underset{(\hat{\bm{x}},\hat{\bm{y}},\hat{\bm{z}}) \sim \mathbb{P}_{\hat{\bm{x}},\hat{\bm{y}},\hat{\bm{z}}}}{\mathbb{E}} \left[(\lvert\lvert \nabla_{\hat{\bm{x}},\hat{\bm{y}},\hat{\bm{z}}} D(\hat{\bm{x}},\hat{\bm{y}},\hat{\bm{z}}) \rvert\rvert_2 - 1)^2\right],
\end{equation}
where points \((\hat{\bm{x}},\hat{\bm{y}},\hat{\bm{z}})\) are drawn at random on straight lines between real or prior samples \((\bm{x},\bm{y},\bm{z})\) and synthesized or inferred samples \((\tilde{\bm{x}},\tilde{\bm{y}},\tilde{\bm{z}})\). The gradient penalty for Jensen-Shannon GAN introduced by \cite{roth17stabilizing} has also been explored, but did not produce better results. The regularization term is set to $\lambda = 10.0$, and $\lambda = 0.01$ for MNIST and SVHN experiments, respectively. 

\subsection{MNIST}
\begin{figure}[!h]
    \centering
    \begin{subfigure}[b]{0.54\textwidth}
        \includegraphics[width=\textwidth]{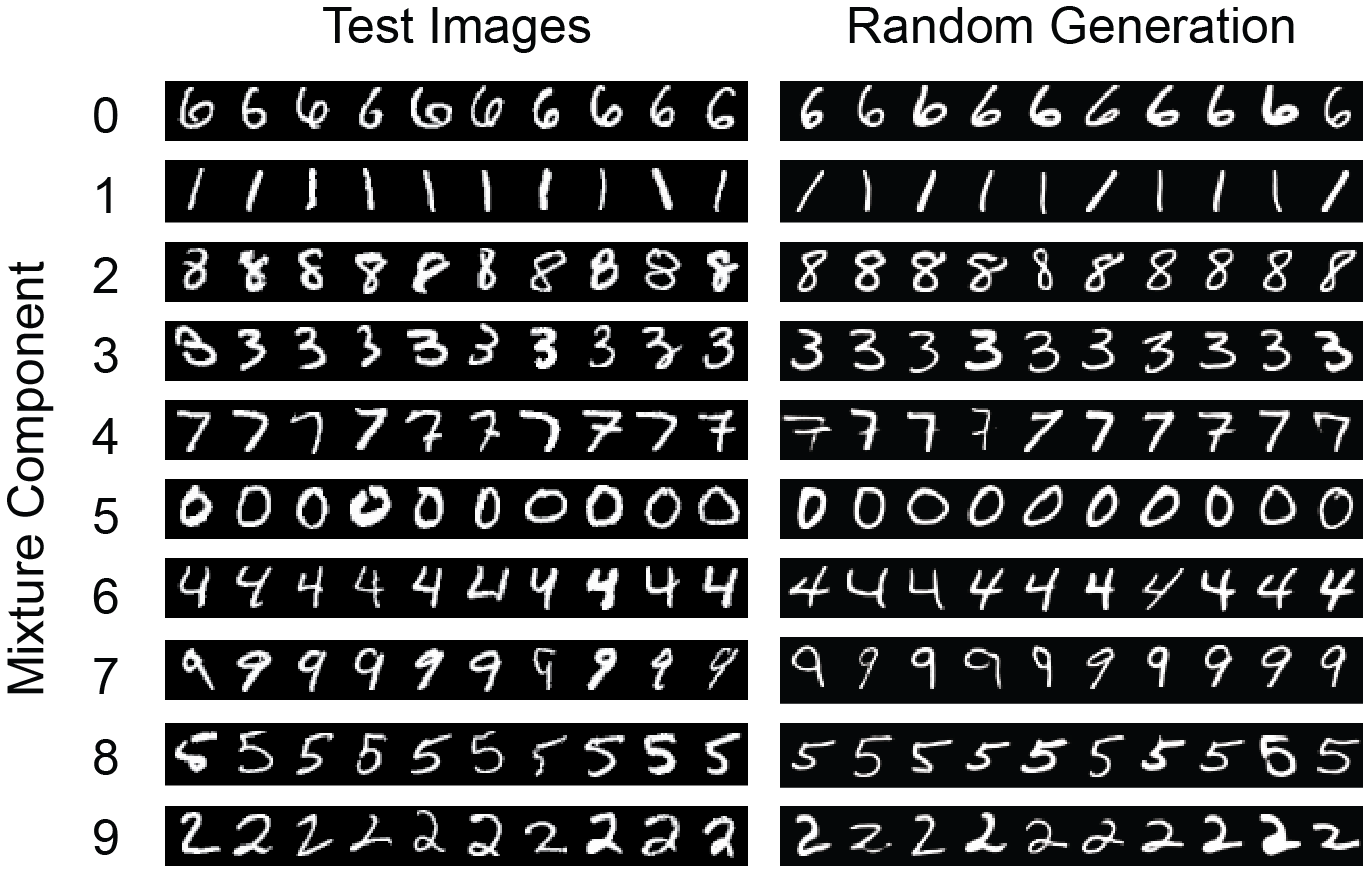}
        \caption{}
        \label{fig:mnist_amm_random}
    \end{subfigure}
    ~
    \begin{subfigure}[b]{0.36\textwidth}
        \includegraphics[width=\textwidth]{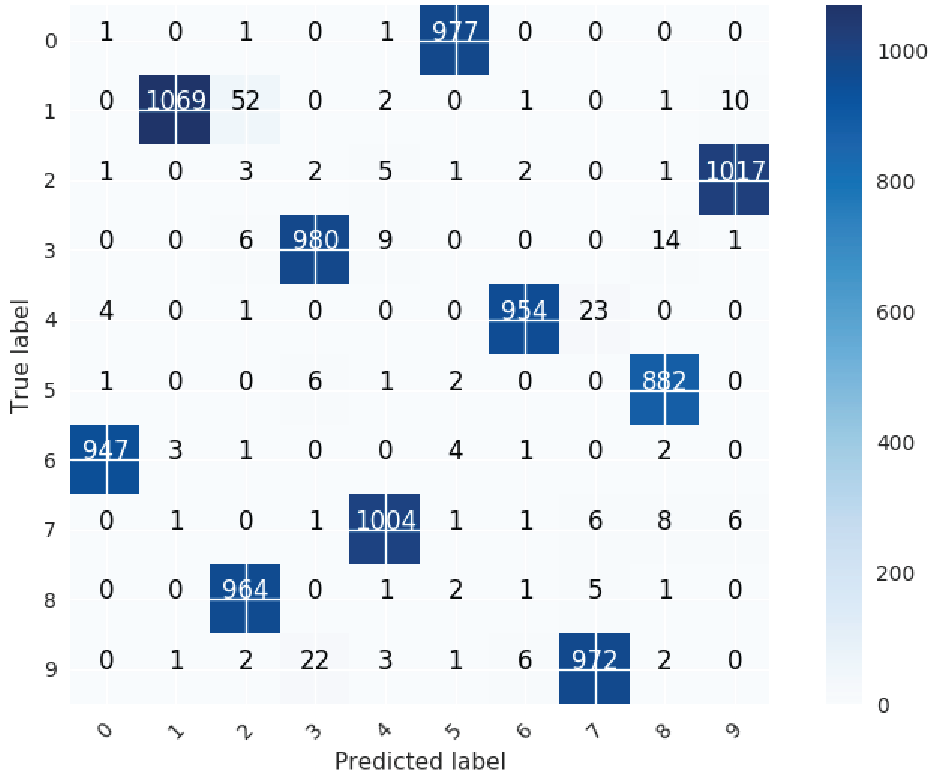}
        \caption{Cluster matrix}
        \label{fig:mnist_amm_cluster}
    \end{subfigure}
    ~
    
    \begin{subfigure}[b]{0.27\textwidth}
        \includegraphics[width=\textwidth]{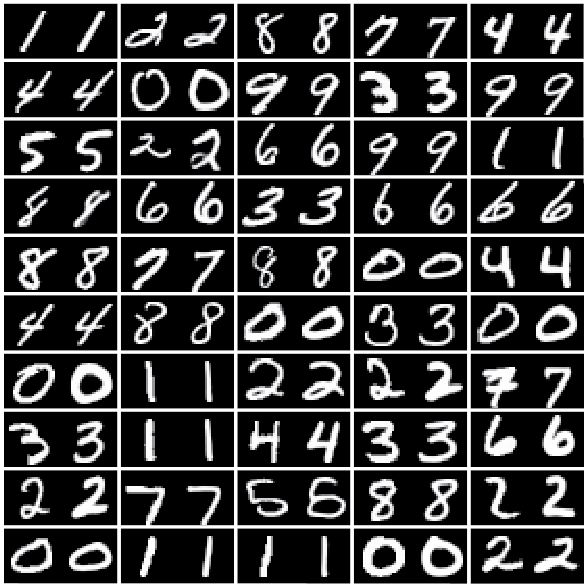}
        \caption{Reconstruction}
        \label{fig:mnist_amm_recon}
    \end{subfigure}
    ~
    \begin{subfigure}[b]{0.27\textwidth}
        \includegraphics[width=\textwidth]{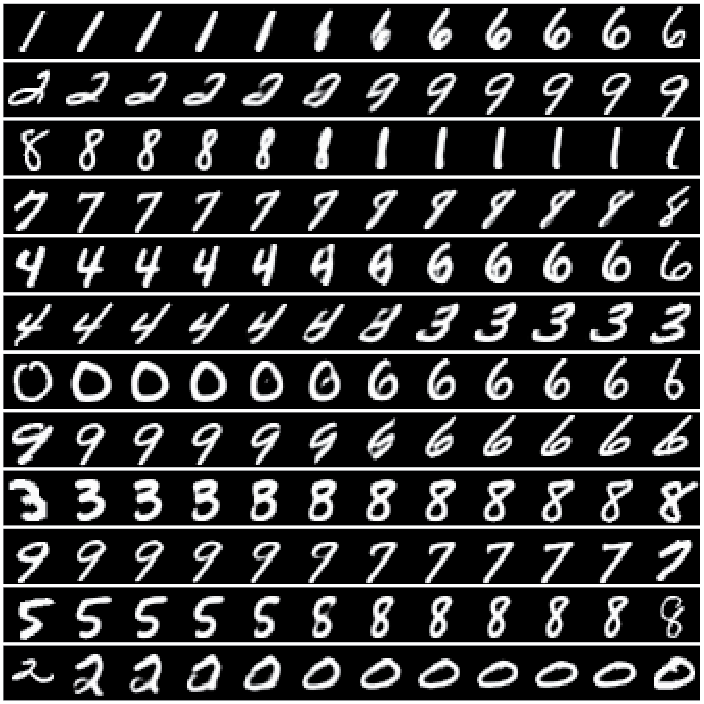}
        \caption{Interpolation}
        \label{fig:mnist_amm_interp}
    \end{subfigure}
    ~
    \begin{subfigure}[b]{0.36\textwidth}
        \includegraphics[width=\textwidth]{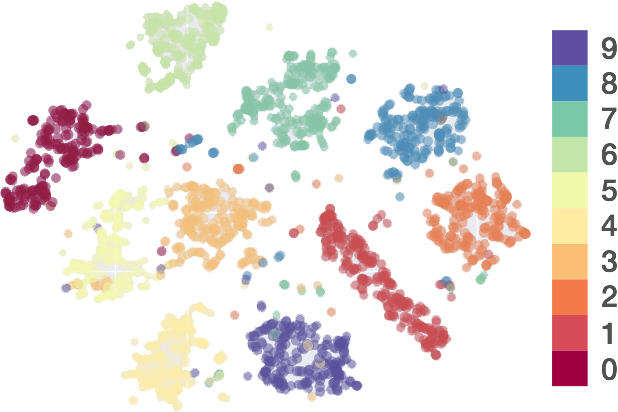}
        \caption{t-SNE projection}
        \label{fig:mnist_amm_tsne}
    \end{subfigure}
    \caption{Unsupervised clustering of MNIST data with 10 mixture components. \emph{(a)} Comparing test image membership and randomly generated digits for each mixture component. \emph{(b)} Cluster matrix: rows correspond to true test labels, and columns correspond to component membership. \emph{(c)} Reconstructions of input images: original data on the left of each pair. \emph{(d)} Interpolation between examples: original data samples are shown in the first and last columns with linearly interpolated generations between. \emph{(e)} t-SNE projection of testing samples, color-coded for the MNIST class labels (0 to 9).}
    \label{fig:mnist_amm}
\end{figure}

In this section, the AMM is evaluated on the task of unsupervised clustering of hand-drawn digits using the MNIST dataset. To model $p(\bm{y})p(\bm{z} \mid \bm{y})$, a 10-component, 64 dimensional mixture of Gaussians is used. A multinomial prior is used for $p(\bm{y})$ with uniform probability for each class. The means of the component distributions are learned using the reparameterization trick via \eqref{eq:z_learned}, and the variance for each distribution is fixed to unit value. Table \ref{tab:unsupervised_mnist} reports the test-set clustering error-rate mean and variance over 5 trials. The AMM achieves $2.86 \pm 0.46$ percent error rate, which is an improvement over the state-of-the-art. Figure \ref{fig:mnist_amm} shows visualizations of results from 1 of the 5 trials.

\begin{table}[!h]
\centering
    \caption{Test set clustering error rate and standard deviation for MNIST data.}\smallskip
\begin{small}
\begin{sc}
    \begin{tabular}{@{}l@{\hskip\tabcolsep}c} \toprule
        Model                                                 & MNIST                      \\ \midrule
        \textbf{CatGAN} \citep{springenberg2015unsupervised}  & $\mathbf{9.70 \pm NR}$  \\
        \textbf{VaDE} \citep{jiang2016variational}            & $\mathbf{5.54 \pm NR}$  \\
        \textbf{InfoGAN} \citep{chen2016infogan}         	  & $\mathbf{5.00 \pm NR}$  \\
        \textbf{AAE} \citep{makhzani2015adversarial}          & $\mathbf{4.10 \pm 1.13}$  \\
        \textbf{AMM}         	                              & $\mathbf{2.86 \pm 0.46}$  \\
        \bottomrule
    \end{tabular}
\end{sc}
\end{small}
\label{tab:unsupervised_mnist}
\end{table}

\subsection{SVHN}
\subsubsection{Unsupervised Clustering}
In this section, unsupervised clustering is revisited. The SVHN dataset is used to investigate how the introduction of confounding attributes, such as color and contrast, affects the semantic separation of digits. To model $p(\bm{y})p(\bm{z} \mid \bm{y})$ a 32 dimensional mixture of 18 spherical, unit variance, Gaussians is used. A multinomial prior is used for $p(\bm{y})$ with uniform probability for each class The means of each distribution are regularly spaced at intervals of 6 units from -6 to 6 along the first two dimensions and from -3 to 3 along the third dimension. The trailing 29 dimensions are set to 0 for each mean. 

\begin{figure}[!h]
    \centering
    \begin{subfigure}[b]{0.40\textwidth}
        \includegraphics[width=\textwidth]{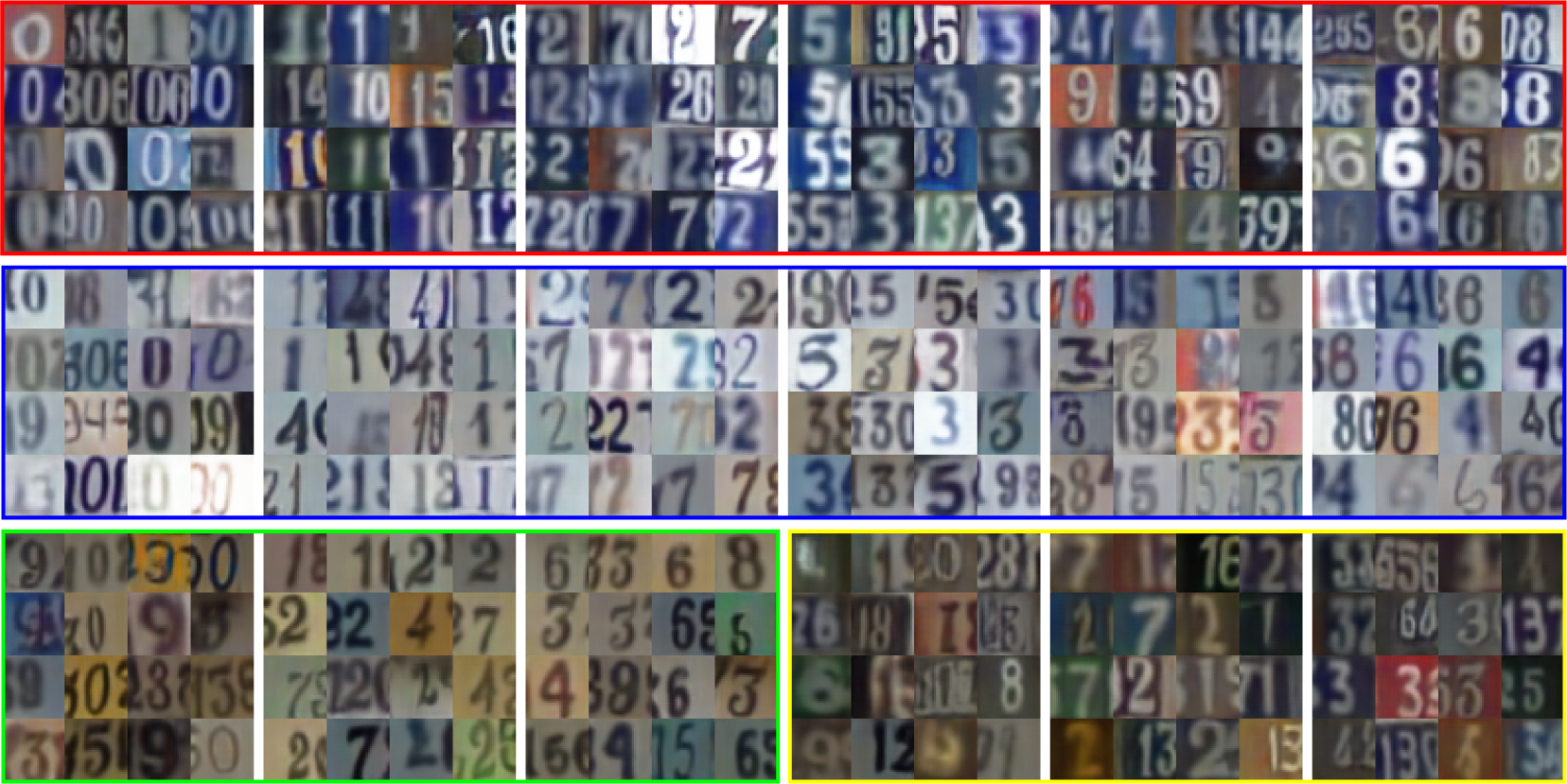}
        \caption{Randomly generated images}
        \label{fig:svhn_amm_random}
    \end{subfigure}
    ~
    \medskip\smallskip
    \begin{subfigure}[b]{0.29\textwidth}
        \includegraphics[width=\textwidth]{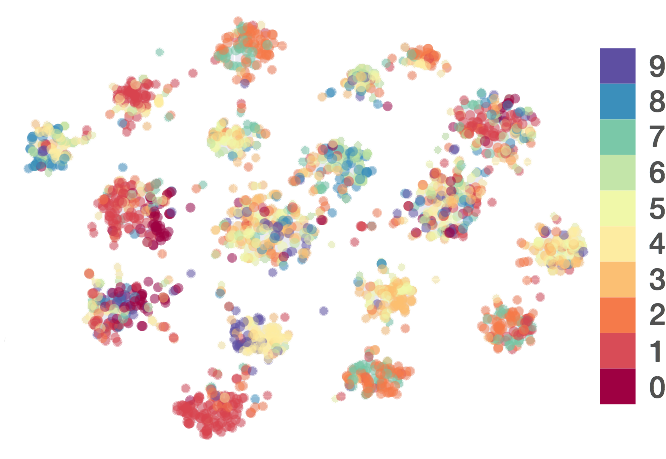}
        \caption{t-SNE projection}
        \label{fig:svhn_amm_tsne}
    \end{subfigure}
    ~
    \begin{subfigure}[b]{0.26\textwidth}
        \includegraphics[width=\textwidth]{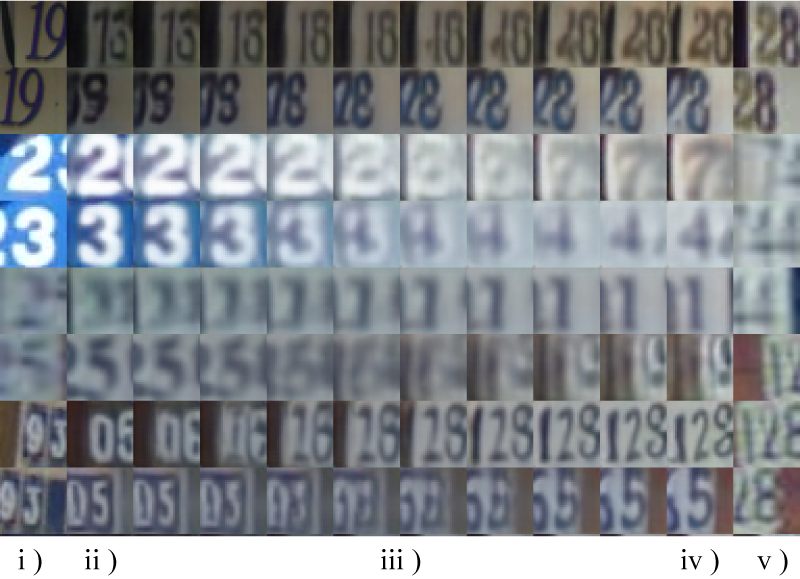}
        \caption{Interpolation}
        \label{fig:svhn_amm_interp}
    \end{subfigure}
    \caption{Unsupervised clustering of SVHN data with 18 mixture components. \emph{(a)} Randomly generated images for each mixture component. The color-boxes delineate four groups of clusters \emph{(Rows 1, 2, 3 (Left) and 3 (Right))} with shared global characteristics. \emph{(b)} t-SNE projection of testing samples, color-coded for the SVHN class label (0 to 9). \emph{(c)} Interpolation between examples: original data samples \emph{(Columns i) and v)}),  associated
reconstructions \emph{(Columns ii) and iv)}), linearly interpolated reconstructions \emph{(Columns iii)}). }
    \label{fig:svhn_amm}
\end{figure}

Figure \ref{fig:svhn_amm_random} shows random samples drawn from each component distribution generated by $G_{\bm{x}}$. We can see four distinct groupings based on the global features of SVHN examples. The top row and last three columns of the bottom row show images with dark backgrounds with light numbers. The middle row and first three columns of the last row show images with light backgrounds and dark numbers. Looking closer at the top two rows we see a nearly symmetric clustering based on number. For example, in the first column we see clusters corresponding to \emph{zero}, in the second column we see clusters corresponding to \emph{one}, and in all of the main groupings we see clusters with numbers \emph{two} and \emph{seven} together. The clusters that combine \emph{twos} and \emph{sevens} are reflected by the orange and green groupings in figure \ref{fig:svhn_amm_tsne}, which is a t-SNE projection of testing samples drawn from $G_{\bm{z}}$ onto a 2D manifold. We show in \ref{fig:svhn_amm_interp} that AMM learns a smooth latent manifold as we interpolate between examples from SVHN.

\subsubsection{Semi-supervised clustering and classification}
It is evident from the last experiment that the confounders introduced by the SVHN dataset made unsupervised semantic clustering more difficult. In this section we show how SAMM can be used to guide clustering along predefined categories using only a small amount of labeled data. To this end we limit the samples drawn from $q(\bm{x}, \bm{y})$ to a random selection of 1000 examples from the training set. To model $p(\bm{y})p(\bm{z} \mid \bm{y})$ we use a 64 dimensional mixture of 10 spherical Gaussians, each with unit variance. In placing the means of each distribution, we take advantage of our prior knowledge of the task. For example, from figure \ref{fig:mnist_amm_tsne}, we can see that \emph{nines} are closer to \emph{fours} than they are to \emph{zeros}, and reflect these assumptions in designing $p(\bm{y})p(\bm{z} \mid \bm{y})$. There is considerable class imbalance in the SVHN dataset so a multinomial prior is used for $p(\bm{y})$ with each class probability set to the frequency observed in the training data. The placement of each mean $\bm{\mu}_k$ within the continuous latent manifold $\bm{z}$ is shown in table \ref{tab:priors} of the appendix. We also run this experiment allowing the $\bm{\mu}_k$'s to be learned using equation \eqref{eq:z_learned}. 

\begin{figure}[!h]
    \centering
    \begin{subfigure}[b]{0.3\textwidth}
    \includegraphics[width=\textwidth]{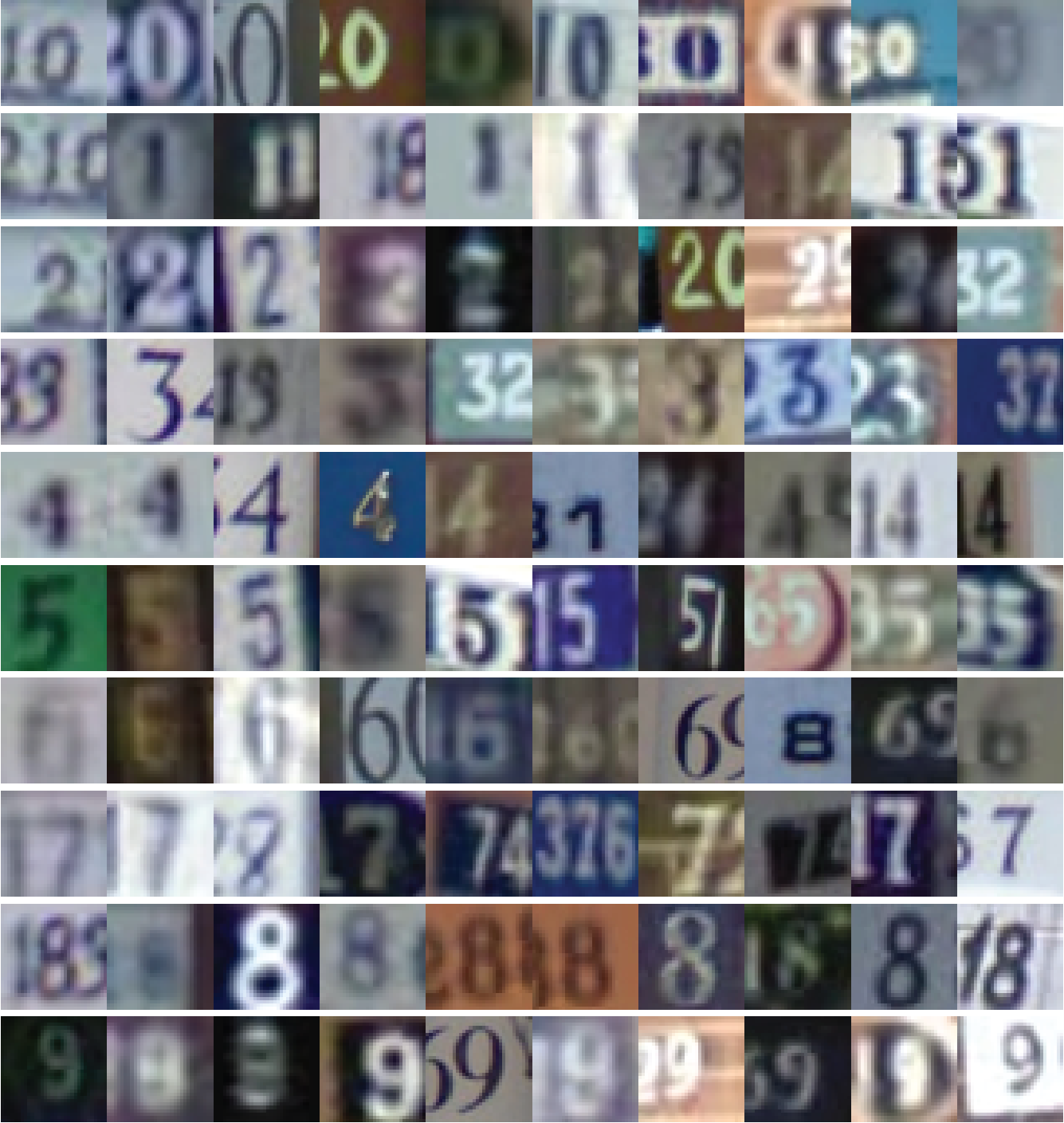}
        \caption{Test images}
        \label{fig:svhn_samm_test}
    \end{subfigure}
    ~
    \begin{subfigure}[b]{0.3\textwidth}
    \includegraphics[width=\textwidth]{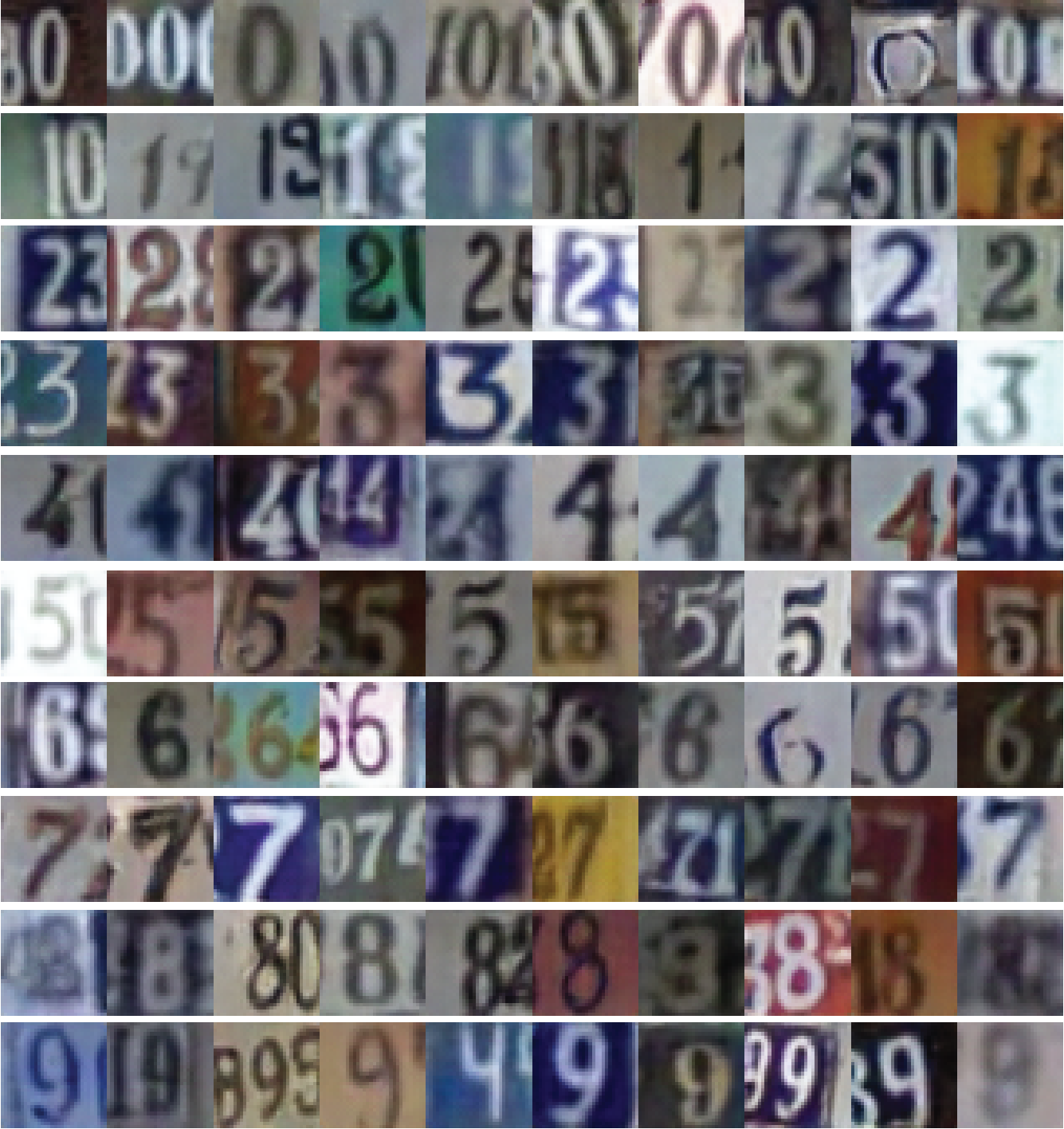}
        \caption{Random}
        \label{fig:svhn_samm_random}
    \end{subfigure}
    ~
    \begin{subfigure}[b]{0.32\textwidth}
        \includegraphics[width=\textwidth]{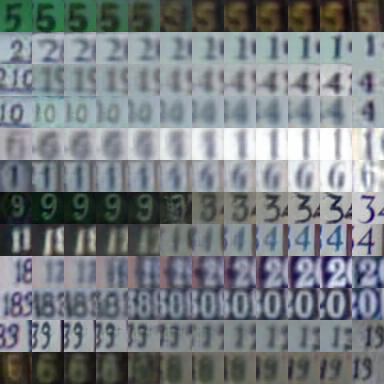}
        \caption{Interpolation}
        \label{fig:svhn_samm_interp}
    \end{subfigure}
    ~
    
    \begin{subfigure}[b]{0.34\textwidth}
        \includegraphics[width=\textwidth]{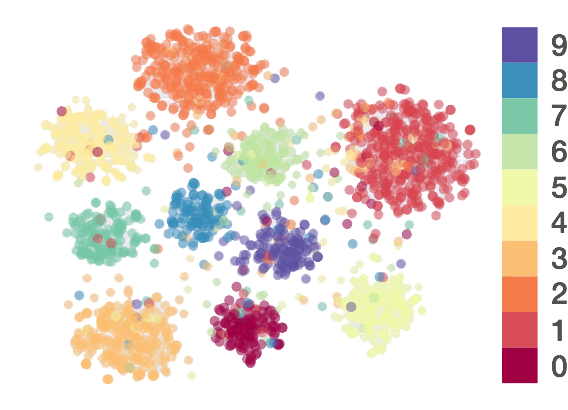}
        \caption{t-SNE projection}
        \label{fig:svhn_samm_tsne}
    \end{subfigure}
    ~
    \begin{subfigure}[b]{0.28\textwidth}
        \includegraphics[width=\textwidth]{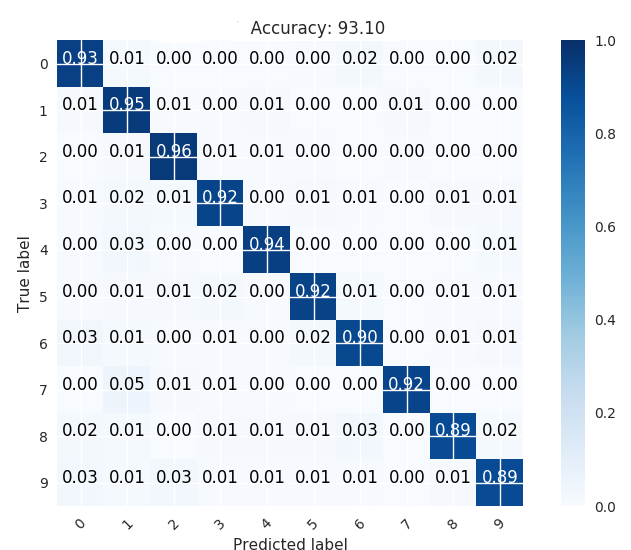}
        \caption{$\tilde{\bm{y}}_{\tilde{\bm{z}}}$}
        \label{fig:svhn_samm_confusion_map}
    \end{subfigure}
    ~
    \begin{subfigure}[b]{0.28\textwidth}
        \includegraphics[width=\textwidth]{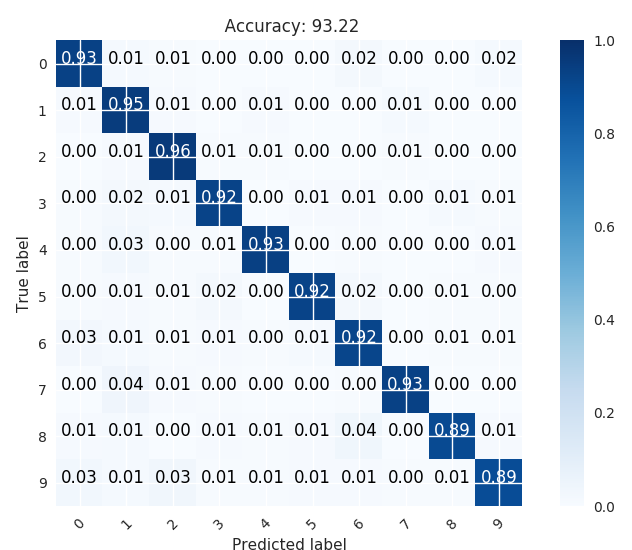}
        \caption{$\tilde{\bm{y}}$}
        \label{fig:svhn_samm_confusion_gy}
    \end{subfigure}
    \caption{Semi-supervised clustering and classification of SVHN data with 10 mixture components. \emph{(a)} Test image predictions: each row corresponds to the predicted class. \emph{(b)} Randomly generated images for each mixture component. \emph{(c)} Interpolation between examples: original data samples in first and last columns. \emph{(d)} t-SNE projection of testing samples, color-coded for the SVHN class label (0 to 9). Confusion matrix for predictions given an image embedding \emph{(e)} and given the generator $G_{\bm{y}}$ \emph{(e)}.}
    \label{fig:svhn_samm}
\end{figure}

Table \ref{tab:semi_supervised_svhn} reports the test-set error-rate mean and variance over 10 trials. SAMM achieves $7.02 \pm 0.17$ percent error rate with the fixed means, and $6.43 \pm 0.12$ when the means are learned, which is an improvement over the ALI baseline. Figure \ref{fig:svhn_samm} shows visualizations of results from 1 of the 10 trials. Finally, given that we have defined $p(\bm{y})p(\bm{z} \mid \bm{y})$ we can use Bayes' theorem to derive $p(\bm{y} \mid \bm{z})$ and get a classifier given an image embedding $\tilde{\bm{z}}$:
\begin{equation}
    \tilde{\bm{y}}_{\tilde{\bm{z}}} = \underset{k}{\mathrm{argmax}} \left[p(\bm{z} = \tilde{\bm{z}} \mid \bm{y} = k)p(\bm{y} = k)\right]
\end{equation}
Figures \ref{fig:svhn_samm_confusion_map} and \ref{fig:svhn_samm_confusion_gy} compare the confusion matrices for predictions given by $\tilde{\bm{y}}_{\tilde{\bm{z}}}$ and those given by $\tilde{\bm{y}}$ from $G_{\bm{y}}$. The similarity between each is further evidence that the inference network has learned to embed data according to the desired distribution.

\begin{table}[!h]
\centering
    \caption{Semi-supervised test set missclassification rate and standard deviation for SVHN data.}\smallskip
\begin{small}
\begin{sc}
    \begin{tabular}{@{}l@{\hskip\tabcolsep}c} \toprule
        Model                                                 & SVHN                      \\ 
                                                              & (\(N = 1000\))            \\ \midrule
        \textbf{AAE} \citep{makhzani2015adversarial}          & $\mathbf{17.70\pm 0.24}$  \\
        \textbf{ImprovedGAN} \citep{salimans2016improved}     & $\mathbf{8.11 \pm 1.30}$  \\
        \textbf{ALI} \citep{dumoulin2016adversarially}        & $\mathbf{7.42 \pm 0.65}$  \\
        \textbf{TripleGAN} \citep{li2017triplegan}            & $\mathbf{5.77 \pm 0.17}$  \\
        \textbf{SGAN} \citep{deng2017structured}         	  & $\mathbf{5.73 \pm 0.12}$  \\ \midrule
        \textbf{SAMM}         	                              & $\mathbf{7.02 \pm 0.17}$  \\
        \textbf{SAMM} Learned $\bm{\mu}_k$    	              & $\mathbf{6.43 \pm 0.12}$  \\
        \bottomrule
    \end{tabular}
\end{sc}
\end{small}
\label{tab:semi_supervised_svhn}
\end{table}

\section{Conclusion}
\label{sec:discussion}

The AMM is presented as a generative model for unsupervised or semi-supervised data clustering. It is the first adversarially optimized method to model the conditional dependence between categorical and continuous latent variables. The AMM achieves state of the art unsupervised clustering results and competitive semi-supervised classification results on benchmark datasets.

\bibliography{references}
\bibliographystyle{references}

\clearpage

\appendix

\section{SAMM Algorithm}
Algorithm \ref{alg:samm} outlines the SAMM training procedure.
\begin{algorithm}
    \begin{algorithmic}
        \State $\theta_{G_{\bm{y}}(\bm{x})}, \theta_{G_{\bm{z}}(\bm{x}, G_{\bm{y}}(\bm{x}))}, \theta_{G_{\bm{z}}(\bm{y})}, \theta_{G_{\bm{x}}(\bm{y}, G_{\bm{z}}(\bm{y}))}, \theta_{D}$ \Comment{Initialize SAMM parameters}
        \While{not done}
            \State $\bm{x}_u^{(1)}, \ldots, \bm{x}_u^{(M)} \sim q(\bm{x})$ \Comment{Sample from unlabeled data and priors}
            \State $\bm{y}_u^{(1)}, \ldots, \bm{y}_u^{(M)} \sim p(\bm{y})$
            \State $\bm{z}_u^{(j)} \sim p(\bm{z} \mid \bm{y} = \bm{y}_u^{(j)}), \quad j = 1, \ldots, M$
            
            \State $\tilde{\bm{x}}_u^{(j)} \sim p(\bm{x} \mid \bm{y} = \bm{y}_u^{(j)}, \bm{z} = \bm{z}_u^{(j)}), \quad j = 1, \ldots, M$ \Comment{Sample from conditionals}
            \State $\tilde{\bm{y}}_u^{(i)} \sim q(\bm{y} \mid \bm{x} = \bm{x}_u^{(i)}), \quad i = 1, \ldots, M$
            \State $\tilde{\bm{z}}_u^{(i)} \sim q(\bm{z} \mid \bm{x} = \bm{x}_u^{(i)}, \bm{y} = \tilde{\bm{y}}_u^{(i)}), \quad i = 1, \ldots, M$
            
            \State $\left(\bm{x}_\ell^{(1)}, \ldots, \bm{x}_\ell^{(M)}\right), \left(\tilde{\bm{y}}_\ell^{(1)}, \ldots, \tilde{\bm{y}}_\ell^{(M)}\right) \sim q(\bm{x}, \bm{y})$ \Comment{Sample from labeled data and priors}
            \State $\bm{y}_\ell^{(1)}, \ldots, \bm{y}_\ell^{(M)} \sim p(\bm{y})$
            \State $\bm{z}_\ell^{(j)} \sim p(\bm{z} \mid \bm{y} = \bm{y}_\ell^{(j)}), \quad j = 1, \ldots, M$
            
            \State $\tilde{\bm{x}}_\ell^{(j)} \sim p(\bm{x} \mid \bm{y} = \bm{y}_\ell^{(j)}, \bm{z} = \bm{z}_\ell^{(j)}), \quad j = 1, \ldots, M$ \Comment{Sample from conditionals}
            \State $\tilde{\bm{z}}_\ell^{(i)} \sim q(\bm{z} \mid \bm{x} = \bm{x}_\ell^{(i)}, \bm{y} = \tilde{\bm{y}}_\ell^{(i)}), \quad i = 1, \ldots, M$
            
            \State $\bm{\rho}_{q_u}^{(i)} \gets D(\bm{x}_u^{(i)}, \tilde{\bm{y}}_u^{(i)}, \tilde{\bm{z}}_u^{(i)}), \quad i = 1, \ldots, M$ \Comment{Compute predictions for unlabeled data}
            \State $\bm{\rho}_{p_u}^{(j)} \gets D(\tilde{\bm{x}}_u^{(j)}, \bm{y}_u^{(j)}, \bm{z}_u^{(j)}), \quad j = 1, \ldots, M$
            
            \State $\bm{\rho}_{q_\ell}^{(i)} \gets D(\bm{x}_\ell^{(i)}, \tilde{\bm{y}}_\ell^{(i)}, \tilde{\bm{z}}_\ell^{(i)}), \quad i =  1, \ldots, M$ \Comment{Compute predictions for labeled data}
            \State $\bm{\rho}_{p_\ell}^{(j)} \gets D(\tilde{\bm{x}}_\ell^{(j)}, \bm{y}_\ell^{(j)}, \bm{z}_\ell^{(j)}), \quad j =  1, \ldots, M$
            
            \State $\mathcal{L}_{D_u} \gets -\frac{1}{2M} \sum_{i=1}^M \log(\bm{\rho}_{q_u}^{(i)}) -\frac{1}{2M} \sum_{j=1}^M\ log(1 - \bm{\rho}_{p_u}^{(j)})$ \Comment{Compute discriminator losses}
            \State $\mathcal{L}_{D_\ell} \gets -\frac{1}{2M} \sum_{i=1}^M \log(\bm{\rho}_{q_\ell}^{(i)}) -\frac{1}{2M} \sum_{j=1}^M\ log(1 - \bm{\rho}_{p_\ell}^{(j)})$
            \State $\mathcal{L}_{G_{\bm{y}_u}(\bm{x})} = \mathcal{L}_{G_{\bm{z}_u}(\bm{x}, G_{\bm{y}}(\bm{x}))} \gets -\frac{1}{2M} \sum_{i=1}^M \log(1-\bm{\rho}_{q_u}^{(i)})$  \Comment{Compute inference losses}
            \State $\mathcal{L}_{G_{\bm{z}_\ell}(\bm{x}, G_{\bm{y}}(\bm{x}))} \gets -\frac{1}{2M} \sum_{i=1}^M \log(1-\bm{\rho}_{q_\ell}^{(i)})$
            \State $\mathcal{L}_{G_{\bm{x}_u}(\bm{y}, G_{\bm{z}}(\bm{y}))} = \mathcal{L}_{G_{\bm{z}_u}(\bm{y})} \gets -\frac{1}{2M} \sum_{i=1}^M \log(\bm{\rho}_{p_u}^{(i)})$ \Comment{Compute $\bm{x}$ generator losses}
            \State $\mathcal{L}_{G_{\bm{x}_\ell}(\bm{y}, G_{\bm{z}}(\bm{y}))} = \mathcal{L}_{G_{\bm{z}_\ell}(\bm{y})} \gets -\frac{1}{2M} \sum_{i=1}^M \log(\bm{\rho}_{p_\ell}^{(i)})$
            
            \State $\theta_{D} \gets \theta_D - \nabla_{\theta_D} \left(\mathcal{L}_{D_u} + \mathcal{L}_{D_\ell}\right)$ \Comment{Update discriminator parameters}
            \State $\theta_{G_{\bm{z}}(\bm{x}, G_{\bm{y}}(\bm{x}))} \gets \theta_{G_{\bm{z}}(\bm{x}, G_{\bm{y}}(\bm{x}))} - \nabla_{\theta_{G_{\bm{z}}(\bm{x}, G_{\bm{y}}(\bm{x}))}} \left(\mathcal{L}_{G_{\bm{z}_u}(\bm{x}, G_{\bm{y}}(\bm{x}))} + \mathcal{L}_{G_{\bm{z}_\ell}(\bm{x}, G_{\bm{y}}(\bm{x}))}\right)$ \Comment{Update $\bm{z}$ inference parameters}
            \State $\theta_{G_{\bm{y}}(\bm{x})} \gets \theta_{G_{\bm{y}}(\bm{x})} - \nabla_{\theta_{G_{\bm{y}}(\bm{x})}} \mathcal{L}_{G_{\bm{y}_u}(\bm{x})}$ \Comment{Update $\bm{y}$ inference parameters}
            \State $\theta_{G_{\bm{x}}(\bm{y}, G_{\bm{z}}(\bm{y}))} \gets \theta_{G_{\bm{x}}(\bm{y}, G_{\bm{z}}(\bm{y}))} - \nabla_{\theta_{G_{\bm{x}}(\bm{y}, G_{\bm{z}}(\bm{y}))}} \left(\mathcal{L}_{G_{\bm{x}_u}(\bm{y}, G_{\bm{z}}(\bm{y}))} + \mathcal{L}_{G_{\bm{x}_\ell}(\bm{y}, G_{\bm{z}}(\bm{y}))}\right)$ \Comment{Update $\bm{x}$ synthesis parameters}
            \State $\theta_{G_{\bm{z}}(\bm{y})} \gets \theta_{G_{\bm{z}}(\bm{y})} - \nabla_{\theta_{G_{\bm{z}}(\bm{y})}} \left(\mathcal{L}_{G_{\bm{z}_u}(\bm{y})} + \mathcal{L}_{G_{\bm{z}_\ell}(\bm{y})}\right)$ \Comment{Update $\bm{z}$ synthesis parameters}
        \EndWhile
    \end{algorithmic}
\caption{SAMM training procedure using distributions (4), (6), and (11).}
\label{alg:samm}
\end{algorithm}

\section{Experiment Information}
\label{sec:model_architecture}
\subsection{Model Architectures}
\begin{figure}[!ht]
    \centering
    \begin{subfigure}[t]{0.72\textwidth}
        \includegraphics[width=\textwidth]{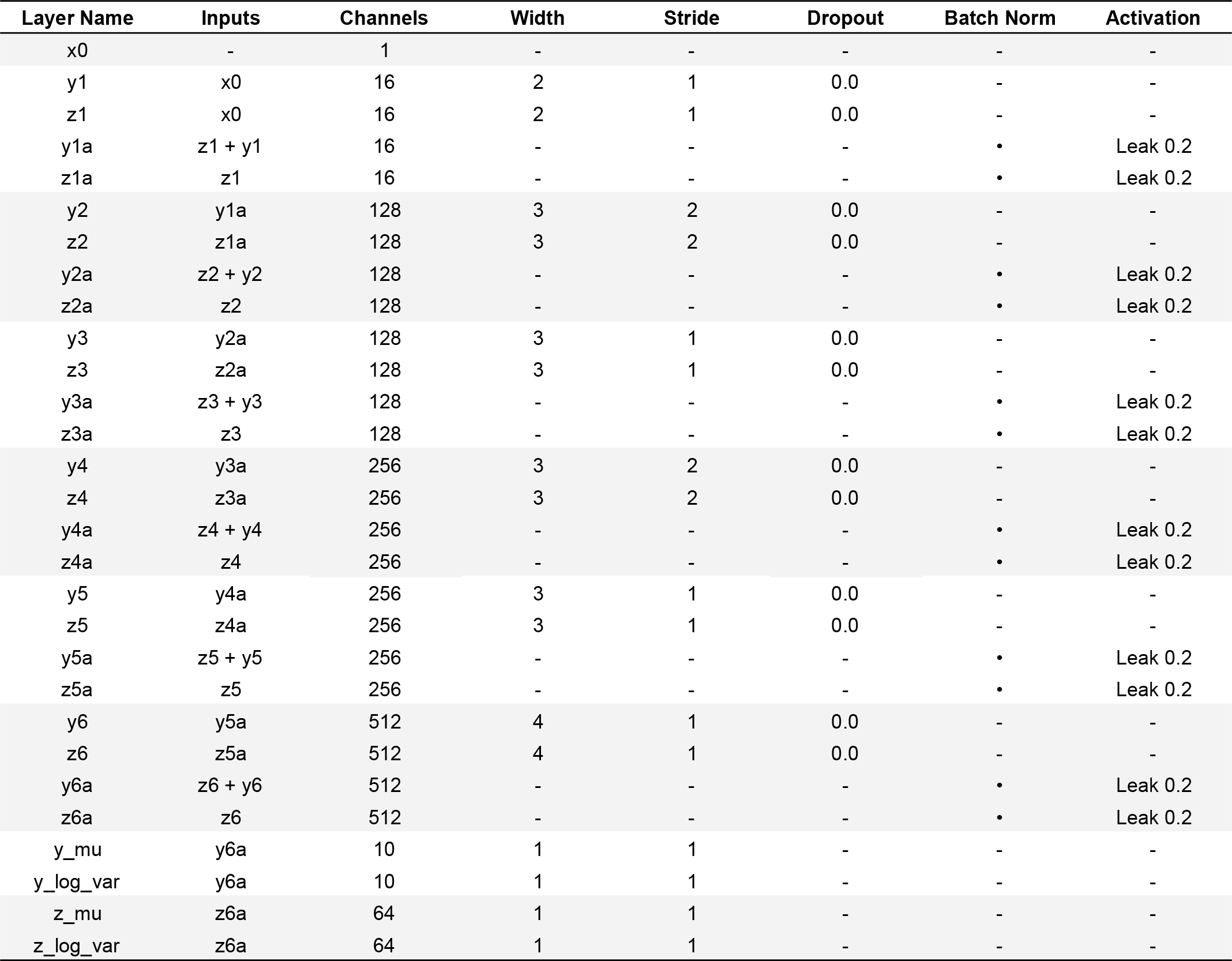}
        \caption{SVHN: $G_{\bm{z}}(\bm{x})G_{\bm{y}}(\bm{x}, G_{\bm{z}}(\bm{x}))$}
        \label{fig:svhn_amm_geny}
    \end{subfigure}
    ~
    
    \begin{subfigure}[t]{0.72\textwidth}
        \includegraphics[width=\textwidth]{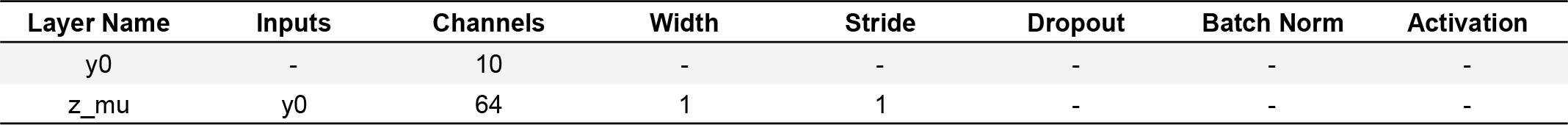}
        \caption{SVHN: $G_{\bm{z}}(\bm{y})$}
        \label{fig:svhn_amm_genz}
    \end{subfigure}
    ~
    
    \begin{subfigure}[t]{0.72\textwidth}
        \includegraphics[width=\textwidth]{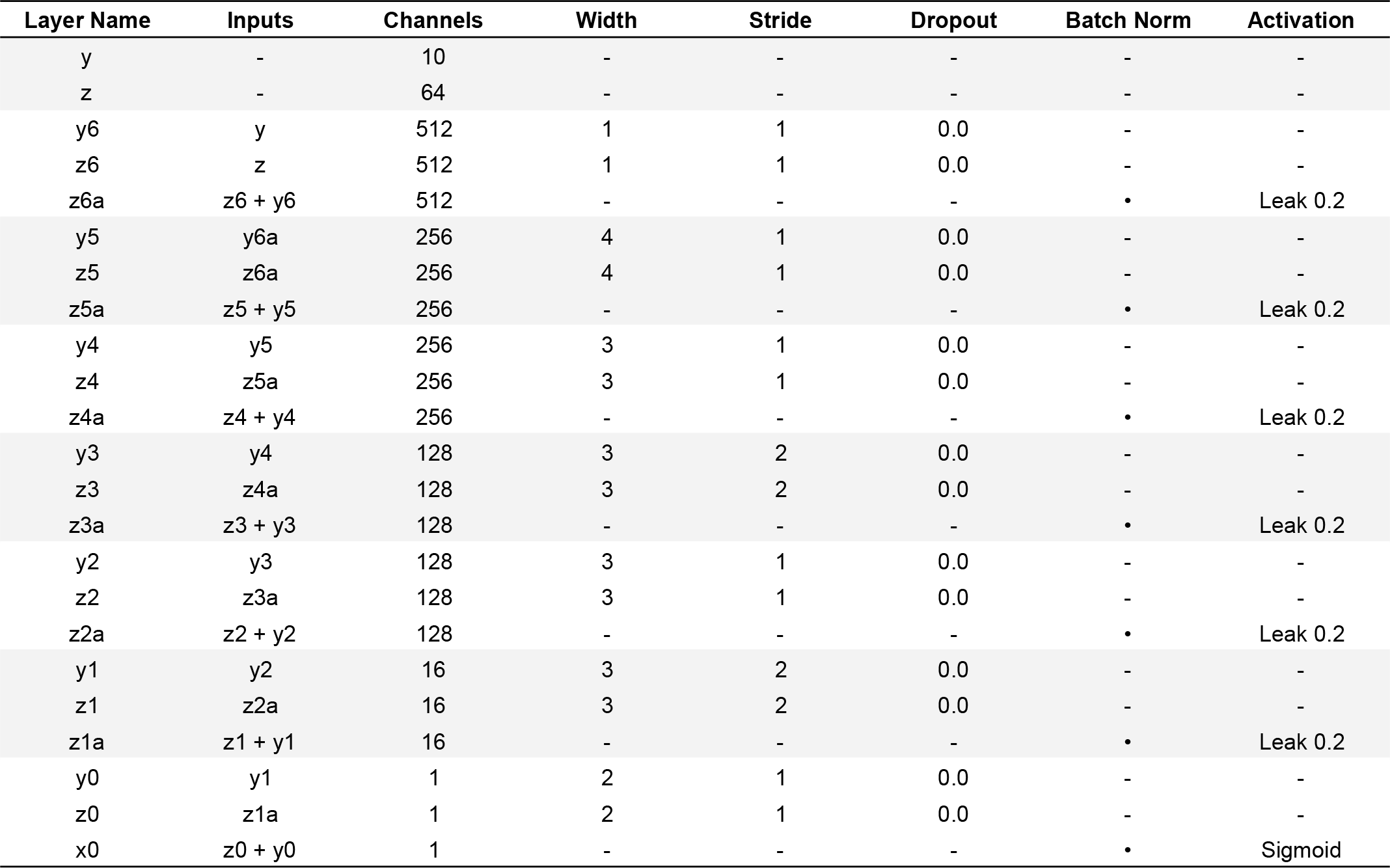}
        \caption{SVHN: $G_{\bm{x}}(\bm{y}, G_{\bm{z}}(\bm{y}))$}
        \label{fig:svhn_amm_genx}
    \end{subfigure}
    ~
    
    \begin{subfigure}[t]{0.72\textwidth}
        \includegraphics[width=\textwidth]{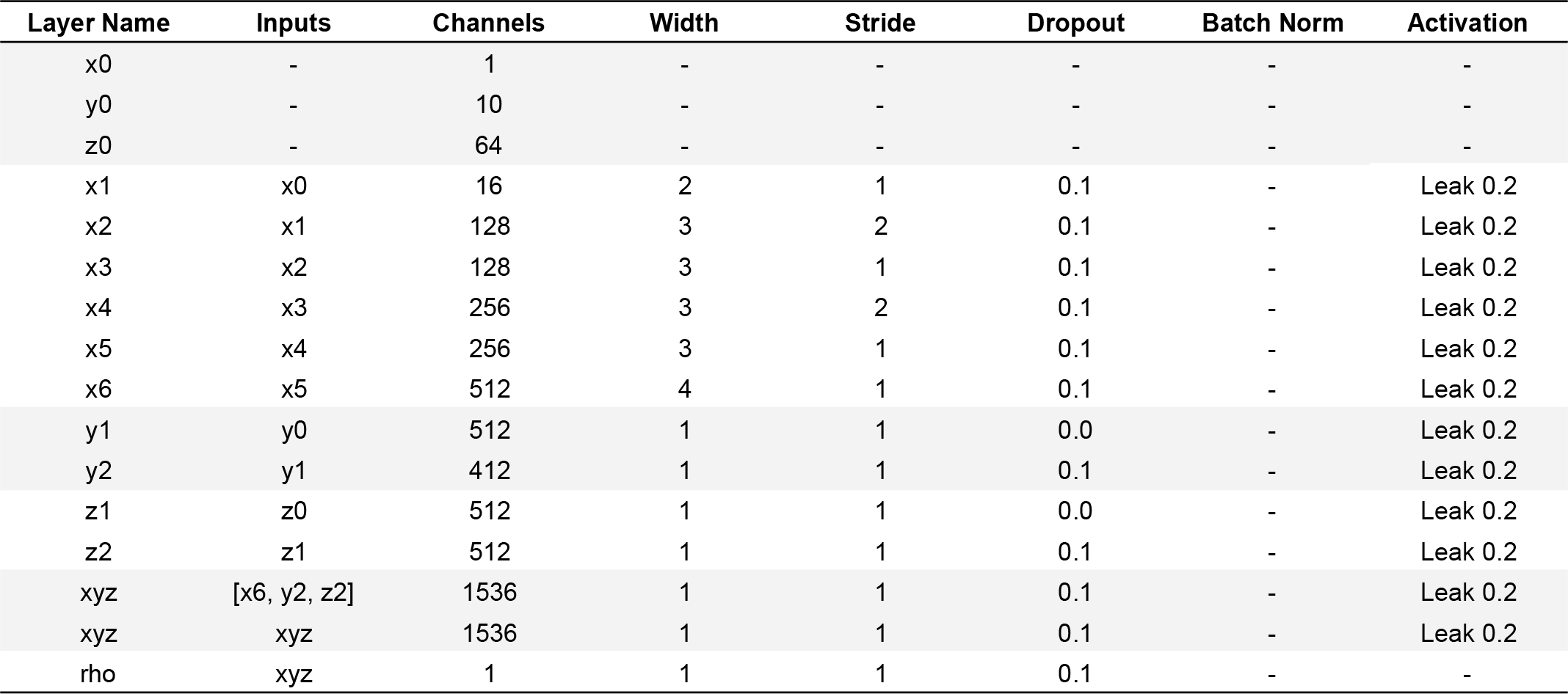}
        \caption{SVHN: $D(\bm{x}, \bm{y}, \bm{z})$}
        \label{fig:svhn_amm_disc}
    \end{subfigure}
    ~
    \caption{Model architecture for SVHN}
    \label{fig:amm_arch_svhn}
\end{figure}

\begin{figure}[!h]
    \centering
    \begin{subfigure}[t]{0.75\textwidth}
        \includegraphics[width=\textwidth]{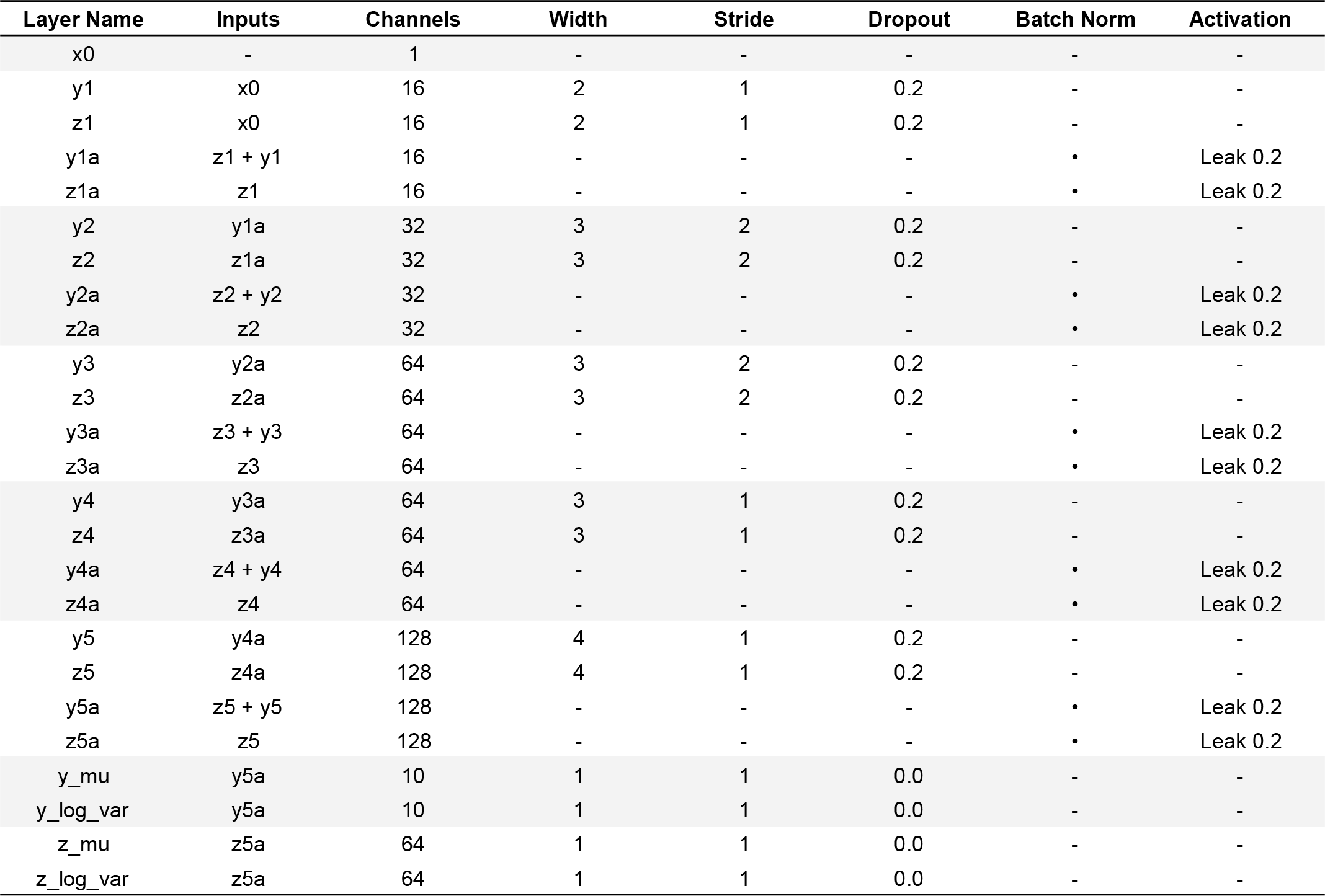}
        \caption{MNIST: $G_{\bm{z}}(\bm{x})G_{\bm{y}}(\bm{x}, G_{\bm{z}}(\bm{x}))$}
        \label{fig:mnist_amm_geny}
    \end{subfigure}
    ~
    
    \begin{subfigure}[t]{0.75\textwidth}
        \includegraphics[width=\textwidth]{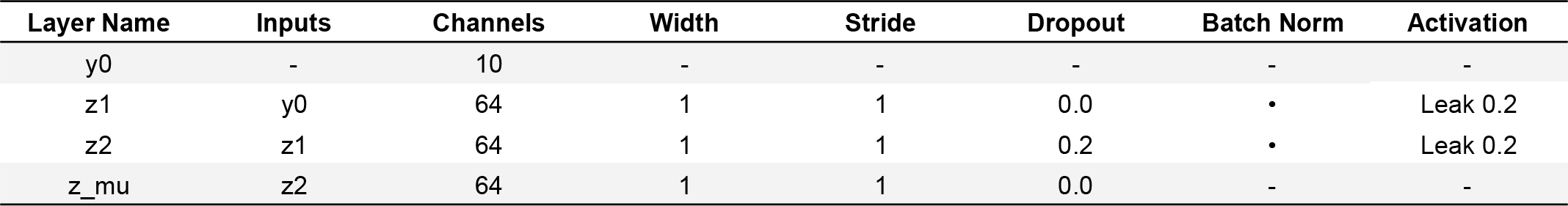}
        \caption{MNIST: $G_{\bm{z}}(\bm{y})$}
        \label{fig:mnist_amm_genz}
    \end{subfigure}
    ~
    
    \begin{subfigure}[t]{0.75\textwidth}
        \includegraphics[width=\textwidth]{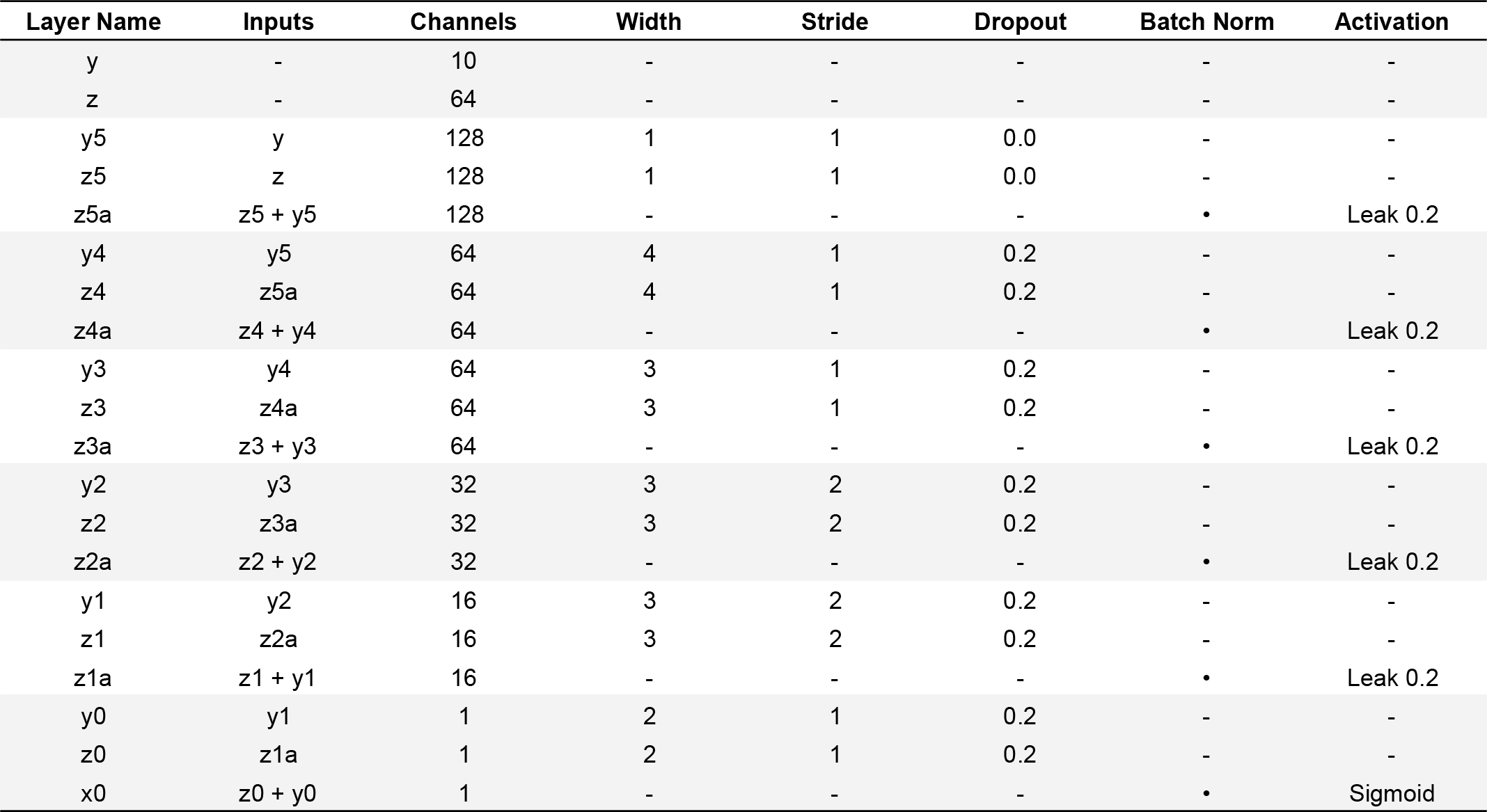}
        \caption{MNIST: $G_{\bm{x}}(\bm{y}, G_{\bm{z}}(\bm{y}))$}
        \label{fig:mnist_amm_genx}
    \end{subfigure}
    ~
    
    \begin{subfigure}[t]{0.75\textwidth}
        \includegraphics[width=\textwidth]{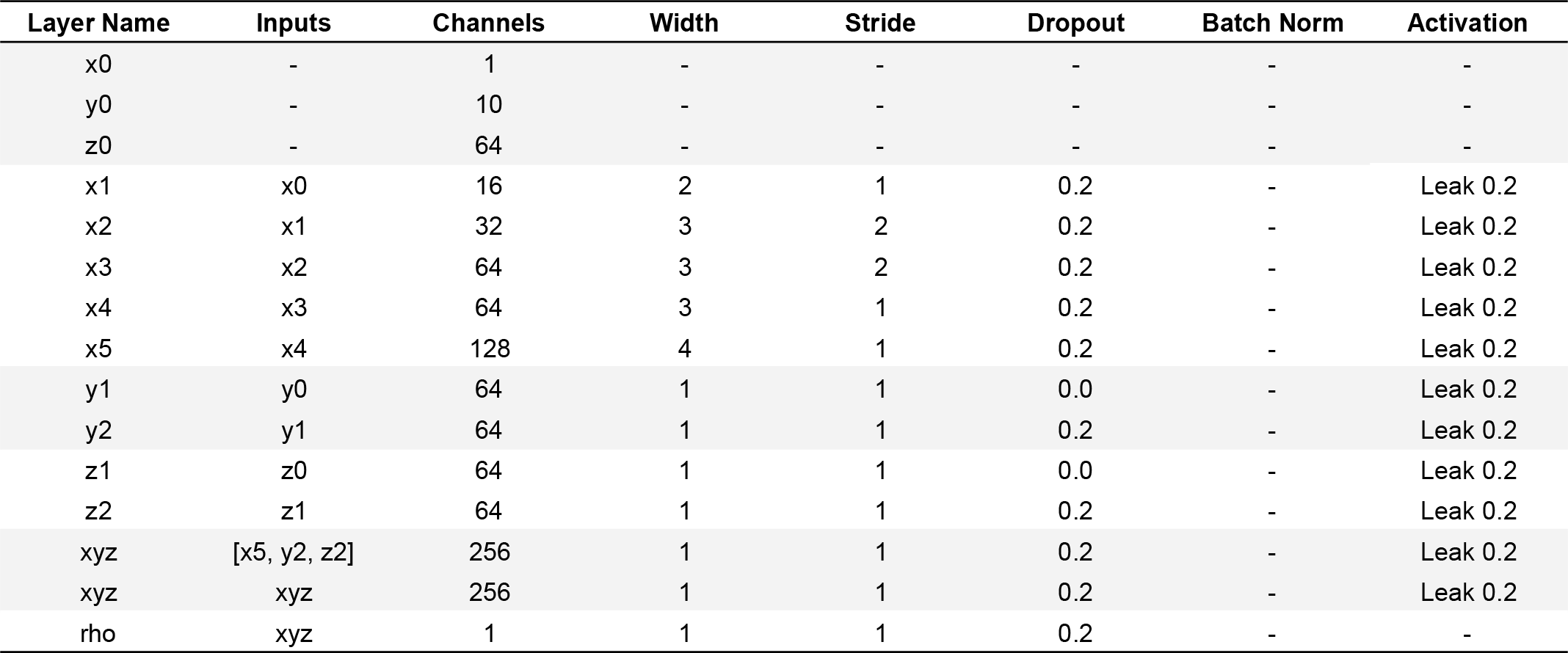}
        \caption{MNIST: $D(\bm{x}, \bm{y}, \bm{z})$}
        \label{fig:mnist_amm_disc}
    \end{subfigure}
    ~
    \caption{Model architecture for MNIST}
    \label{fig:amm_arch}
\end{figure}

Figures \ref{fig:amm_arch_svhn} and \ref{fig:amm_arch} detail the model architectures for the SVHN and MNIST experiments, respectively.

\subsection{Mean Placement}
The placement of each mean for the fixed mean semi-supervised SVHN experiment is shown in table \ref{tab:priors}

\begin{table}[!h]
    \centering
    \caption{SVHN Semi-Supervised: Placement of means for $p(\bm{y})p(\bm{z} \mid \bm{y})$}\smallskip
    \begin{small}
    \begin{sc}
    \begin{tabular}{cccccc}
    \toprule
        Mean  & $z_0$  & $z_1$  & $z_2$  & $z_3$  & $z_{4-31}$ \\ 
        \midrule
        $\mu_0$       & -3       & 3        & -3       & -3       & 0  \\
        $\mu_1$       & -3       & -3       &  3       &  3       & 0  \\
        $\mu_2$       & -3       & 3        &  3       & -3       & 0  \\
        $\mu_3$       &  3       & -3       & -3       & -3       & 0  \\
        $\mu_4$       & -3       & -3       &  3       & -3       & 0  \\
        $\mu_5$       &  3       & -3       &  3       & -3       & 0  \\
        $\mu_6$       &  3       & 3        &  3       & -3       & 0  \\
        $\mu_7$       & -3       & 3        &  3       &  3       & 0  \\
        $\mu_8$       &  3       & 3        & -3       & -3       & 0  \\
        $\mu_9$       & -3       & -3       & -3       & -3       & 0  \\
        \bottomrule
    \end{tabular}
    \end{sc}
    \end{small}
    \label{tab:priors}
\end{table}

\end{document}